\documentclass{article}

\usepackage{microtype}
\usepackage{graphicx}
\usepackage{subfigure}
\usepackage{booktabs} 

\usepackage{hyperref}

\usepackage[accepted]{icml2024}

\usepackage{amsmath}
\usepackage{amssymb}
\usepackage{mathtools}
\usepackage{amsthm}

\usepackage[toc,page,header]{appendix}
\usepackage{minitoc}

\usepackage{float}

\usepackage[capitalize,noabbrev]{cleveref}

\usepackage[capitalize]{cleveref}
\crefname{appendix}{App.}{Apps.}
\crefname{section}{Sec.}{Secs.}
\Crefname{section}{Sec.}{Sections}
\Crefname{table}{Table}{Tables}
\crefname{table}{Tab.}{Tabs.}
\Crefname{figure}{Fig.}{Figs.}
\Crefname{equation}{Eq.}{Eqs.}

\usepackage{amsmath,amsfonts,bm,mathtools}

\def\Figref#1{Figure~\ref{#1}}

\def\eqref#1{equation~\ref{#1}}

\def\1{\bm{1}}

\def\rvd{{\mathbf{d}}}

\def\rvn{{\mathbf{n}}}

\def\rvx{{\mathbf{x}}}
\def\rvy{{\mathbf{y}}}
\def\rvz{{\mathbf{z}}}

\def\mA{{\bm{A}}}

\def\mD{{\bm{D}}}
\def\mE{{\bm{E}}}

\def\mI{{\bm{I}}}

\DeclareMathAlphabet{\mathsfit}{\encodingdefault}{\sfdefault}{m}{sl}
\SetMathAlphabet{\mathsfit}{bold}{\encodingdefault}{\sfdefault}{bx}{n}

\def\gB{{\mathcal{B}}}

\def\gE{{\mathcal{E}}}
\def\gF{{\mathcal{F}}}
\def\gG{{\mathcal{G}}}
\def\gH{{\mathcal{H}}}

\def\gN{{\mathcal{N}}}

\newcommand{\E}{\mathbb{E}}

\newcommand{\R}{\mathbb{R}}

\theoremstyle{plain}

\theoremstyle{definition}

\theoremstyle{remark}

\usepackage[textsize=tiny]{todonotes}

\def\setstretch#1{\renewcommand{\baselinestretch}{#1}}
\icmltitlerunning{DisCo-Diff: Enhancing Continuous Diffusion Models with Discrete Latents}

\begin{document}

\twocolumn[
\icmltitle{\textit{DisCo-Diff:} Enhancing Continuous Diffusion Models with Discrete Latents}



\icmlsetsymbol{equal}{*}

\begin{icmlauthorlist}
\icmlauthor{Yilun Xu}{equal,nv,mit}
\icmlauthor{Gabriele Corso}{mit}
\icmlauthor{Tommi Jaakkola}{mit}
\icmlauthor{Arash Vahdat}{nv}
\icmlauthor{Karsten Kreis}{nv}

\centerline{\href{https://research.nvidia.com/labs/lpr/disco-diff/}{https://research.nvidia.com/labs/lpr/disco-diff}}
\end{icmlauthorlist}

\icmlaffiliation{mit}{MIT}
\icmlaffiliation{nv}{NVIDIA}

\icmlcorrespondingauthor{Yilun Xu}{yilunx@nvidia.com}

\icmlkeywords{Machine Learning, ICML}

\vskip 0.3in
]



\printAffiliationsAndNotice{\icmlEqualContribution} 

\doparttoc 
\faketableofcontents 

\vspace{
-9mm
}
\begin{abstract}
\vspace{
4mm
}
Diffusion models (DMs) have revolutionized generative learning. They utilize a diffusion process to encode data into a simple Gaussian distribution. However, encoding a complex, potentially multimodal data distribution into a single \textit{continuous} Gaussian distribution arguably represents an unnecessarily challenging learning problem. We propose \textit{\textbf{Dis}crete-\textbf{Co}ntinuous Latent Variable \textbf{Diff}usion Models (DisCo-Diff)} to simplify this task by introducing complementary \textit{discrete} latent variables. We augment DMs with learnable discrete latents, inferred with an encoder, and train DM and encoder end-to-end. 
DisCo-Diff does not rely on pre-trained networks, making the framework universally applicable. The discrete latents significantly simplify learning the DM's complex noise-to-data mapping by reducing the curvature of the DM's generative ODE.
An additional autoregressive transformer models the distribution of the discrete latents, a simple step because DisCo-Diff requires only few discrete variables with small codebooks. 
We validate DisCo-Diff on toy data, several image synthesis tasks as well as molecular docking, and find that introducing discrete latents consistently improves model performance. For example, DisCo-Diff achieves state-of-the-art FID scores on class-conditioned ImageNet-64/128 datasets with ODE sampler.
\end{abstract}

\begin{figure*}[t!]
  \begin{minipage}[c]{0.63\textwidth}
    \includegraphics[width=1.0\textwidth]{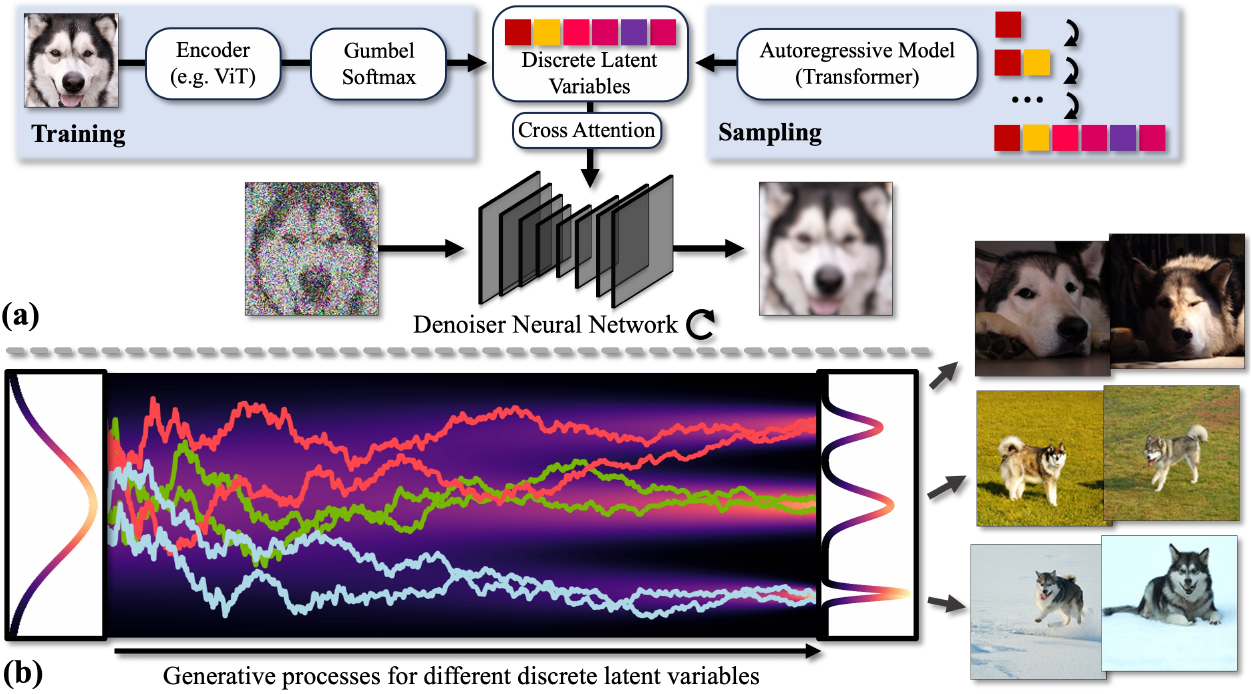}
  \end{minipage}\hfill
  \begin{minipage}[c]{0.36\textwidth}
  \vspace{-3mm}
    \caption{\small \textbf{Dis}crete-\textbf{Co}ntinuous Latent Variable \textbf{Diff}usion Models (DisCo-Diff) augment DMs with additional \textit{discrete} latent variables that capture global appearance patterns, here shown for images of huskies. \textbf{(a)} During training, discrete latents are inferred through an encoder, for images a vision transformer~\cite{dosovitskiy2021vit}, and fed to the DM via cross-attention. Backpropagation is facilitated by continuous relaxation with a Gumbel-Softmax distribution. To sample novel images, an additional autoregressive model is learnt over the distribution of discrete latents. \textbf{(b)} Schematic visualization of generative denoising diffusion trajectories. Different colors indicate different discrete latent variables, pushing the trajectories toward different modes.}
    \label{fig:pipeline}
  \end{minipage}
  \vspace{-4mm}
\end{figure*}

\section{Introduction} \label{sec:intro}
Diffusion models (DMs)~\cite{sohl2015deep,ho2020ddpm,song2020score} have recently led to breakthroughs for generative modeling in diverse domains. For instance, they can synthesize expressive high-resolution imagery~\cite{saharia2022photorealistic,ramesh2022dalle2,rombach2021highresolution,balaji2022ediffi} or they can generate accurate molecular structures~\cite{corso2023diffdock,yim2023framediff,ingraham2022chroma,watson2023rfdiffusion}. DMs leverage a forward diffusion process that effectively encodes the training data in a simple, unimodal Gaussian prior distribution.
Generation can be formulated either as a stochastic or, more conveniently, as a deterministic process that takes as input random noise from the Gaussian prior and transforms it into data through a generative ordinary differential equation (ODE)~\cite{song2020score}.
The Gaussian prior corresponds to the DM's \textit{continuous latent variables}, where the data is uniquely encoded through the ODE-defined mapping.

However, realistic data distributions are typically high-dimensional, complex and often multimodal. Directly encoding such data into a single unimodal Gaussian distribution and learning a corresponding reverse noise-to-data mapping is challenging. The mapping, or generative ODE, necessarily needs to be highly complex, with strong curvature, and one may consider it unnatural to map an entire data distribution to a single Gaussian distribution. 
In practice, conditioning information, such as class labels or text prompts, often helps to simplify the complex mapping by offering the DM's denoiser additional cues for more accurate denoising. However, such conditioning information is typically of a semantic nature and, even given a class or text prompt, the mapping remains highly complex. For instance, in the case of images, even within a class we find images with vastly different styles and color patterns, which corresponds to large distances in pixel space.

Here, we propose \textit{\textbf{Dis}crete-\textbf{Co}ntinuous Latent Variable \textbf{Diff}usion Models (DisCo-Diff)}, DMs augmented with additional \textit{discrete} latent variables that encode additional high-level information about the data and can be used by the main DM to simplify its denoising task (\Cref{fig:pipeline}). These discrete latents are inferred through an encoder network
and learnt end-to-end together with the DM.
Thereby, the discrete latents directly learn to encode information that is beneficial for reducing the DM's score matching objective and making the DM's hard task of mapping simple noise to complex data easier. Indeed, in practice, we find that they significantly reduce the curvature of the DM's generative ODE and reduce the DM training loss in particular for large diffusion times, where denoising is most ambiguous and challenging.
In contrast to previous work~\cite{bao2022conditional,hu2023SelfGuidedDM,harvey2023visual}, we do not rely on domain-specific pre-trained encoder networks, making our framework general and universally applicable. To facilitate sampling of discrete latent variables during inference, we learn an autoregressive model over the discrete latents in a second step. We only use a small set of \textit{discrete} latents with relatively small codebooks, which makes the additional training of the autoregressive model easy. We specifically advocate for the use of auxiliary discrete instead of continuous latents; see \Cref{sec:disco_motivation}.\looseness=-1

While previous works \cite{esser2021vqgan,ramesh2021dalle,chang2022maskgit,yu2022scaling,pernias2023wuerstchen,chang2023muse} use fully discrete latent variable-based approaches to model images, this typically requires large sets of spatially arranged latents with large codebooks, which makes learning their distribution challenging. 
DisCo-Diff, in contrast, carefully combines its discrete latents with the continuous latents (Gaussian prior) of the DM and effectively separates the modeling of discrete and continuous variations within the data. It requires only a few discrete latents.\looseness=-1

To demonstrate its universality, we validate the DisCo-Diff framework on several different tasks. As a motivating example, we study 2D toy distributions, where the discrete latents learn to capture different modes with smaller curvature during sampling. 
We then tackle image synthesis, where the discrete latents learn large-scale appearance, often associated with global style and color patterns. Thereby, they offer complementary benefits to semantic conditioning information.
Quantitatively, DisCo-Diff universally boosts output quality and achieves state-of-the-art performance on several ImageNet generation benchmarks.
In addition, we experimentally validate that auxiliary \textit{discrete} latents are superior to \textit{continuous} latents in our setup, and study different network architectures for injecting the discrete latents into the DM network. A careful hierarchical design can encourage different discrete latents to encode different image characteristics, such as shape vs. color, reminiscent of observations from the literature on generative adversarial networks~\cite{karras2019stylegan,karras2020stylegan2}.
We also apply DisCo-Diff to molecular docking, a critical task in drug discovery, where the discrete latents again improve performance by learning to indicate critical atoms in the interaction and, in this way, deconvolving the multimodal uncertainty given by different possible poses from continuous variability of each pose. Moreover, we augment Poisson Flow Generative Models~\cite{xu2022pfgm,xu2023pfgmpp} with discrete latent variables to showcase that the framework can also be applied to other ``iterative'' generative models, other than regular DMs, observing similar benefits.\looseness=-1


\textbf{Contributions.}
\textit{(i)} We propose DisCo-Diff, a novel framework for combining discrete and continuous latent variables in DMs in a universal manner.
\textit{(ii)} We extensively validate DisCo-Diff, significantly boosting model quality in all experiments, 
and achieving state-of-the-art performance on several image synthesis tasks.
\textit{(iii)} We present detailed analyses as well as ablation and architecture design studies that demonstrate the unique benefits of discrete latent variables and how they can be fed to the main denoiser network.
\textit{(iv)} Overall, we provide insights for designing performant generative models. We make the case for discrete latents by showing 
that real-world data is best modeled with generative frameworks that leverage \textit{both} discrete and continuous latents.
We intentionally developed a simple and universal framework that does not rely on pre-trained encoders to offer a broadly applicable modeling approach to the community.


\section{Background}
DisCo-Diff builds on (continuous-time) DMs~\cite{song2020score}, and we follow the EDM framework~\cite{Karras2022ElucidatingTD}. DMs perturb the clean data {$\rvy \sim p_{\textrm{data}}(\rvx)$} in a fixed forward process using $\sigma^2(t)$-variance Gaussian noise, where $\rvy \in \R^d$ and $t$ denotes the time along the diffusion process. The resulting distribution is denoted as $p(\rvx; \sigma(t))$ with $\rvx \in \R^d$.
For sufficiently large $\sigma_\textrm{max}$, this distribution is almost identical to pure random Gaussian noise. DMs leverage this observation to sample $\rvx_0\sim \mathcal{N}(\rvx_0; \mathbf{0},\sigma_\textrm{max}^2\mI)$ and then iteratively denoise the sample through a sequence of $M+1$ gradually decreasing noise levels $\sigma_{i+1}<\sigma_{i} \,(\sigma_0=\sigma_\textrm{max})$, where $i\in[0,...,M]$ and $\rvx_i \sim p(\rvx; \sigma_i)$. The $\sigma_i$ correspond to a discretization of a continuous $\sigma(t)$ function. If $\sigma_M=0$, then the final $\rvx_M$ follows the data distribution.
Sampling corresponds to simulating a deterministic or stochastic differential equation
\begin{equation}\label{eq:generation_sde}
\begin{split}
    & d\rvx = \underbrace{-\dot{\sigma}(t)\sigma(t)\boldsymbol{\nabla}_\rvx \log p(\rvx; \sigma(t)) dt}_{\textrm{Probability Flow ODE}} \\
    &\underbrace{- \beta(t)\sigma^2(t) \boldsymbol{\nabla}_\rvx \log p(\rvx; \sigma(t)) dt + \sqrt{2\beta(t)}\sigma(t)d\boldsymbol{\omega}_t}_{\textrm{Langevin Diffusion SDE}},
\end{split}
\end{equation}
where $d\boldsymbol{\omega}_t$ is a standard Wiener process and $\boldsymbol{\nabla}_\rvx \log p(\rvx; \sigma(t))$ is the \textit{score function} of the diffused distribution $p(\rvx; \sigma(t))$. The first term in \Cref{eq:generation_sde} is the Probability Flow ODE, which pushes samples from large to small noise levels. The second term is a Langevin Diffusion SDE, an equilibrium sampler for different noise levels $\sigma(t)$, which can help correct errors during synthesis~\cite{Karras2022ElucidatingTD}. This component can be scaled by the time-dependent parameter $\beta(t)$. Setting $\beta(t)=0$ leads to pure ODE-based synthesis.
Generally, different sampling methods can be used to solve the generative ODE/SDE.

Training a DM corresponds to learning a model to approximate the intractable score function $\boldsymbol{\nabla}_\rvx \log p(\rvx; \sigma(t))$. Following the EDM framework, we parametrize $\boldsymbol{\nabla}_\rvx \log p(\rvx; \sigma(t))= (\mD_\theta(\rvx, \sigma(t))-\rvx)/\sigma^2(t)$, where $\mD_\theta(\rvx, \sigma(t))$ is a learnable \textit{denoiser} neural network that is trained to predict clean data from noisy inputs and is conditioned on the noise level $\sigma(t)$. It can be trained using denoising score matching~\cite{hyvarinen2005scorematching,lyu2009scorematching,vincent2011,song2019generative}, minimizing\looseness=-1
\begin{equation}
    \mathbb{E}_{\rvy\sim p_{\textrm{data}}(\rvy)} \mathbb{E}_{t,\rvn}\left[\lambda(t)||\mD_\theta(\rvy+\rvn, \sigma(t))-\rvy||^2\right]
    \label{eq:dsm}
\end{equation}
where $t\sim p(t)$ for a distribution $p(t)$ over diffusion times $t$, $\rvn\sim \mathcal{N}(\rvn;\mathbf{0},\sigma^2(t)\mI)$, and $\lambda(t)$ is a function that gives different weight to the objective for different noise levels.

In this work, we use $\sigma(t)=t$ and follow the EDM work's configuration~\cite{Karras2022ElucidatingTD}, unless otherwise noted. Moreover, we also leverage classifier-free guidance in DisCo-Diff when conditioning on the discrete latent variables. Classifier-free guidance combines the score functions of an unconditional and a conditional diffusion model to amplify the conditioning; see~\citet{ho2021classifierfree}.
\begin{figure}[t]
    \centering
    \subfigure[$128\times 128$]{\includegraphics[width=0.23\textwidth]{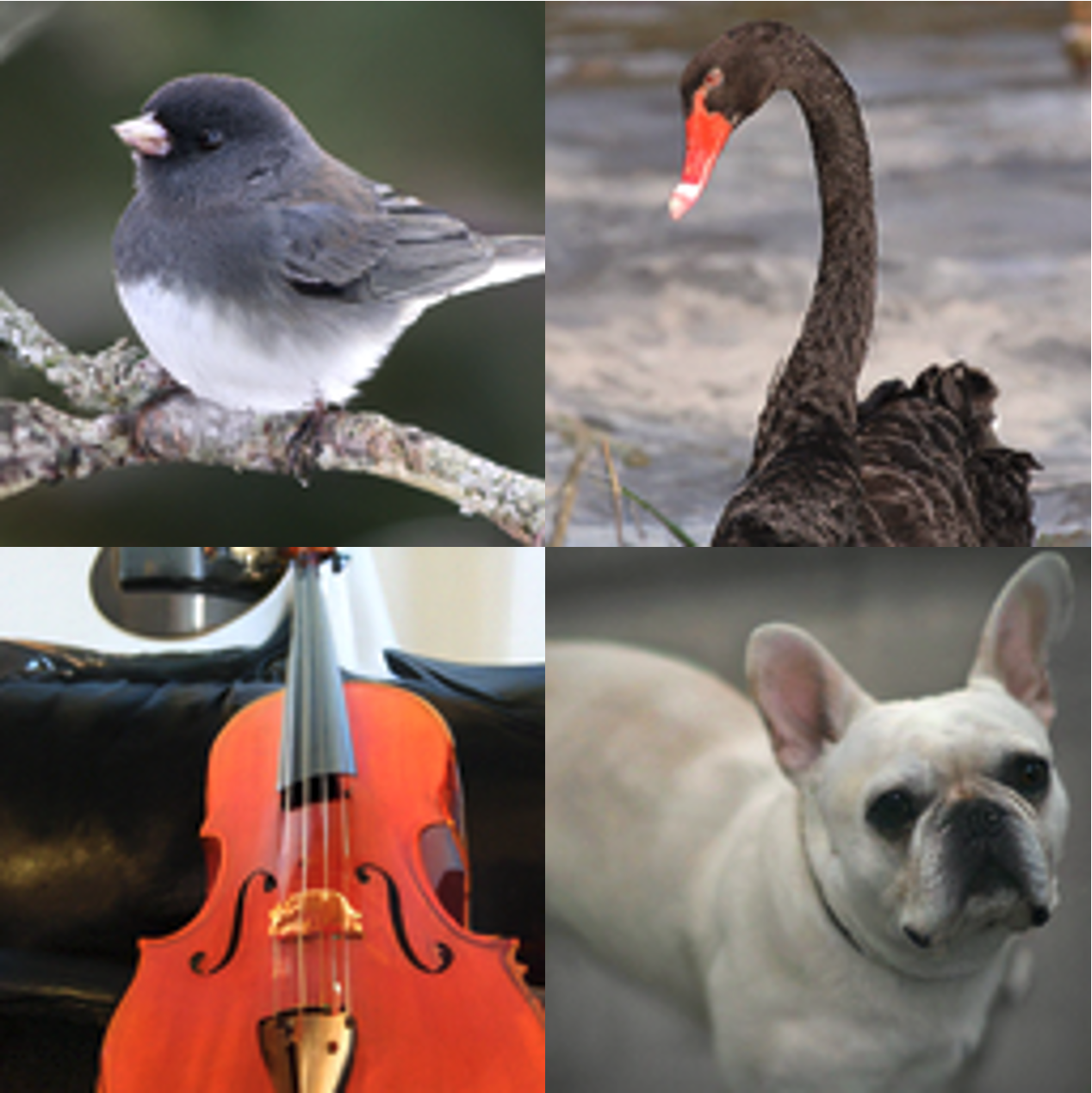}}
    \subfigure[Shared discrete latents]{\includegraphics[width=0.23\textwidth, trim=0cm 0.38cm 0cm 0cm]{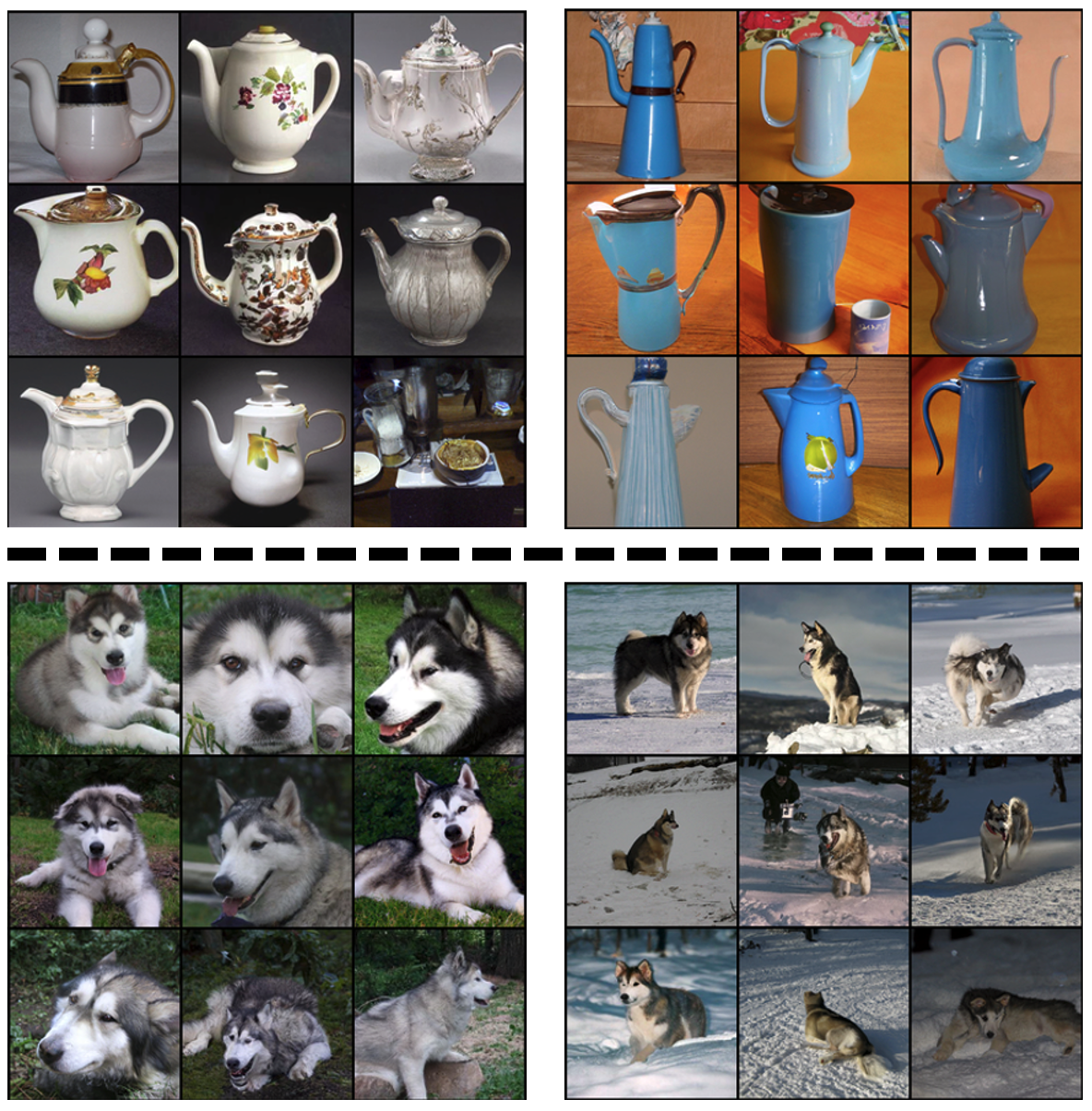}}
    \vspace{-4mm}
    \caption{Samples generated from DisCo-Diff trained on the ImageNet dataset: (a) randomly sampled discrete latents and class labels; (b) samples in each grid sharing the same discrete latent. The class label for the top/bottom row is fixed to coffeepot/malamute.}
    \label{fig:vis-128}
    \vspace{-4mm}
\end{figure}
\vspace{-2mm}
\section{DisCo-Diff}\label{sec:method}
\vspace{-1mm}
In \Cref{sec:disco_obj}, we first formally define DisCo-Diff's generative model and training framework, before discussing and carefully motivating our approach in detail in \Cref{sec:disco_motivation}. In \Cref{sec:disco_arch}, we highlight critical architecture considerations.

\subsection{Generative Model and Training Objective}\label{sec:disco_obj}
In our DisCo-Diff framework (\Cref{fig:pipeline}), we augment a DM's learning process with an $m$-dimensional discrete latent $\rvz \in \mathbb{N}^m$, where each dimension is a random variable from a categorical distribution of codebook size $k$. There are three learnable components: the denoiser neural network $\mD_\theta \colon\R^d\times \R \times \mathbb{N}^m \to\R^d$, corresponding to DisCo-Diff's DM, which predicts denoised images conditioned on diffusion time $t$ and discrete latent $\rvz$; an encoder $\mE_\phi\colon\R^d \to \mathbb{N}^m$, used to infer discrete latents given clean images $\rvy$. It outputs a categorical distribution over the $k$ categories for each discrete latent; and a post-hoc auto-regressive model $\mA_\psi$, which approximates the distribution of the learned discrete latents $\rvz$ by $\prod_{i=1}^m p_\psi(\rvz_i | \rvz_{<i})$. DisCo-Diff's training process is divided into two stages. In the first stage, the denoiser $\mD_\theta$ and the encoder $\mE_\phi$ are co-optimized in an end-to-end fashion. This is achieved by extending the denoising score matching objective (as expressed in Eq.~\ref{eq:dsm}) to include learnable discrete latents $\rvz$ associated with each data $\rvy$: \looseness=-1
\begin{equation}
    \mathbb{E}_{\rvy} \mathbb{E}_{\rvz \sim \mE_\phi(\rvy)} \mathbb{E}_{t,\rvn}\left[\lambda(t)||\mD_\theta(\rvy+\rvn, \sigma(t), \rvz)-\rvy||^2\right],\label{eq:obj-disco}
\end{equation}
where $\rvy\sim p_{\textrm{data}}(\rvy)$. In contrast to the standard objective in Eq.~\ref{eq:dsm}, which focuses on learning the reparameterization of the score $\boldsymbol{\nabla}_\rvx \log p(\rvx; \sigma(t))$, the denoiser in our approach is essentially learning the reparameterization of the conditional score $\boldsymbol{\nabla}_\rvx \log p(\rvx|\rvz; \sigma(t))$, with the convolution of the probability density functions $p(\cdot|\rvz; \sigma(t)) = p(\cdot|\rvz) * \mathcal{N}(\mathbf{0},\sigma^2(t)\mI)$. This conditional score originates from conditioning the DM on the discrete latents $\rvz$, which are inferred by the encoder $\mE_\phi$. The denoiser network $\mD_\theta$ can better capture the time-dependent score~(\textit{i.e.,} achieving a reduced loss) if the score for each sub-distribution $p(\rvx|\rvz; \sigma(t))$ is simplified. Therefore, the encoder $\mE_\phi$, which has access to clean input data, is encouraged to encode useful information into the discrete latents and help the denoiser to more accurately reconstruct the data.
Naively backpropagating gradients into the encoder through the sampling of the discrete latent variables $\rvz$ is not possible. Hence, during training we rely on a continuous relaxation based on the Gumbel-Softmax distribution~\cite{Jang2016CategoricalRW} (see \Cref{app:imagenet_exps_details} for details).\looseness=-1

When training the denoiser network, we randomly replace the discrete latent variables with a non-informative null-embedding with probability $0.1$. Thereby, the DM learns both a discrete latent variable-conditioned and a regular, unconditional score. During sampling, we can combine these scores for classifier-free guidance~\cite{ho2021classifierfree} with respect to the model's own discrete latents, and amplify their conditioning effect (details in \Cref{app:cfg}).

We can interpret DisCo-Diff as a variational autoencoder (VAE)~\cite{kingma2014vae,rezende2014stochastic,vandenoord2017vqvae,rolfe2017discrete} with discrete latents and a DM as decoder. VAEs often employ regularization on their latents. We did not find this to be necessary, as
we use only very low-dimensional latent variables, \textit{e.g.,} 10 in our ImageNet experiments, with relatively small codebooks. Moreover, we employ a strictly non-zero temperature in the Gumbel-Softmax relaxation, encouraging stochasticity.

In the second stage, we train the autoregressive model $\mA_\psi$ to capture the distribution of the discrete latent variables $p_\phi(\rvz)$ defined by pushing the clean data through the trained encoder.
We use a maximum likelihood objective as follows:
\begin{equation}
     \mathbb{E}_{\rvy\sim p_{\textrm{data}}(\rvy), \rvz \sim \mE_\phi(\rvy)} \left[\sum_{i=1}^m \log p_\psi(\rvz_i|\rvz_{<i})\right] \label{eq:obj-ar}
\end{equation}
Since we set $m$ to a relatively small number,
it becomes very easy for the model to handle such short discrete vectors, which makes this second-stage training efficient. Also the additional sampling overhead due to this autoregressive component on top of the DM becomes negligible. At inference time, when using DisCo-Diff to generate novel samples, we first sample a discrete latent variable from the autoregressive model, and then sample the DM with an ODE or SDE solver. We provide the algorithm pseudocode for training and sampling in Appendix~\ref{app:algo}.


\begin{figure}[t!]
    \centering
   \includegraphics[width=0.5\textwidth]{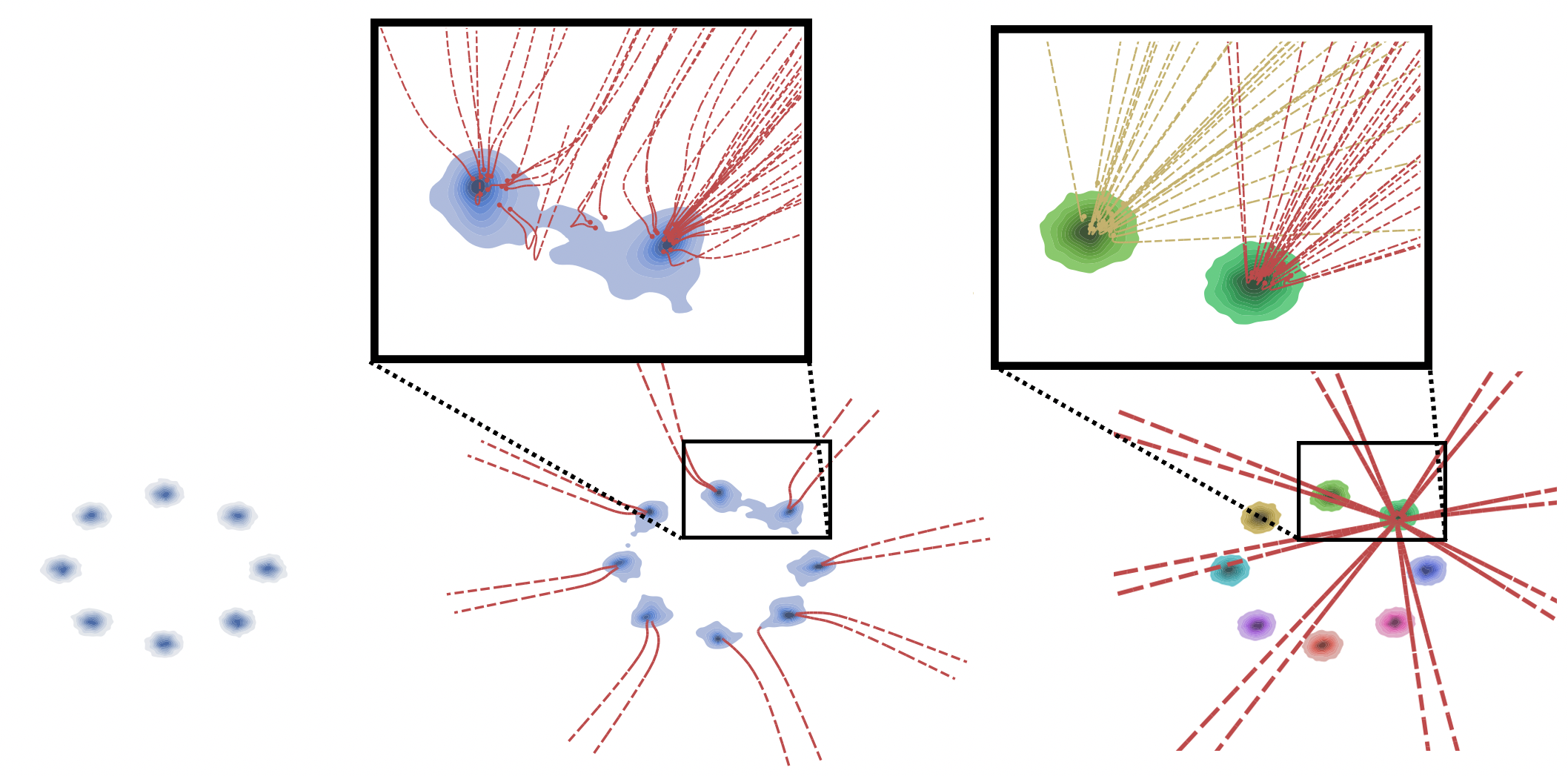}
   \vspace{-9mm}
    \caption{\small \textbf{Modeling 2D mixture of Gaussians.} \textit{Left:} Data distribution. \textit{Middle:} Generated data by regular DM. \textit{Right:} Generated data by DisCo-Diff. We use different colors to distinguish data generated by different discrete latents. We further provide zoom-ins and visualize some ODE trajectories by dotted lines.}
    \label{fig:toy-vis}
    \vspace{-4mm}
\end{figure}

\vspace{-2mm}
\subsection{Motivation and Related Work}\label{sec:disco_motivation}
\vspace{-1mm}
We will now critically discuss and motivate our design choices and also discuss the most relevant related works. \textit{For an extended discussion of related work see \Cref{app:related_work}.}

\textbf{The curvature of diffusion models.} DMs, in their simpler ODE-based formulation ($\beta(t)=0$ in \Cref{eq:generation_sde}), learn a complex noise-to-data mapping. The noise is drawn from an analytically tractable, unimodal Gaussian distribution. As the data is encoded in this distribution, we can consider this high-dimensional Gaussian distribution the DM's \textit{continuous} latent variables (DMs can generally be seen as deep latent variable models~\cite{huang2021perspective,kingma2021variational}). However, the mapping from unstructured noise to a diverse, typically multimodal data distribution necessarily needs to be highly complex. This corresponds to a highly non-linear generative ODE with strong curvature, which is challenging to learn and also makes synthesis slow by requiring a fine discretization. To illustrate this point, we trained a DM on a simple 2D mixture of Gaussians, where we observe bent ODE trajectories near the data (\Cref{fig:toy-vis}, \textit{middle}). This effect is significantly stronger in high dimensions.

\begin{figure}[t!]
    \centering
   \subfigure{\includegraphics[width=0.235\textwidth]{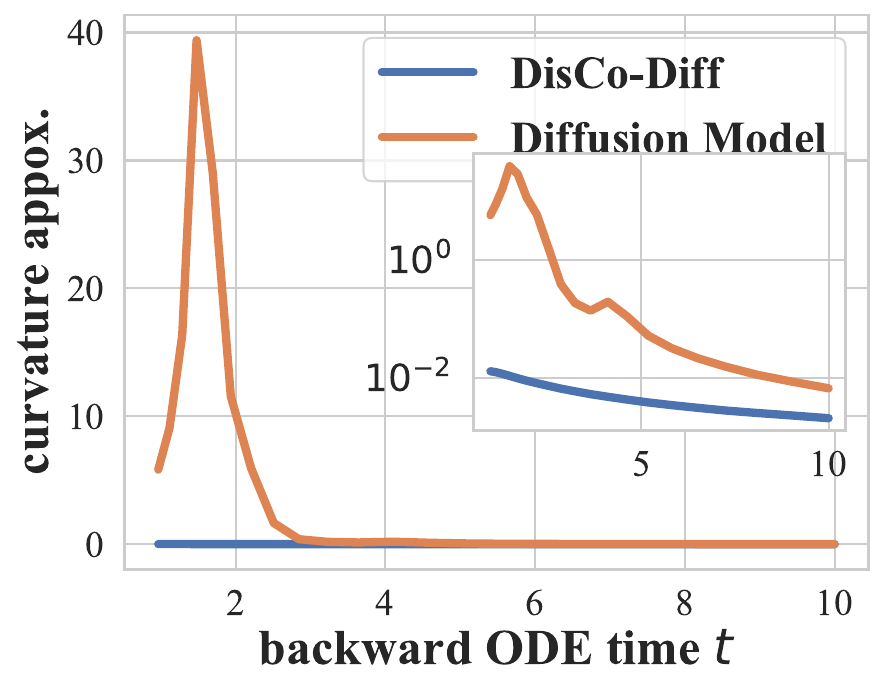}}
      \subfigure{\includegraphics[width=0.235\textwidth]{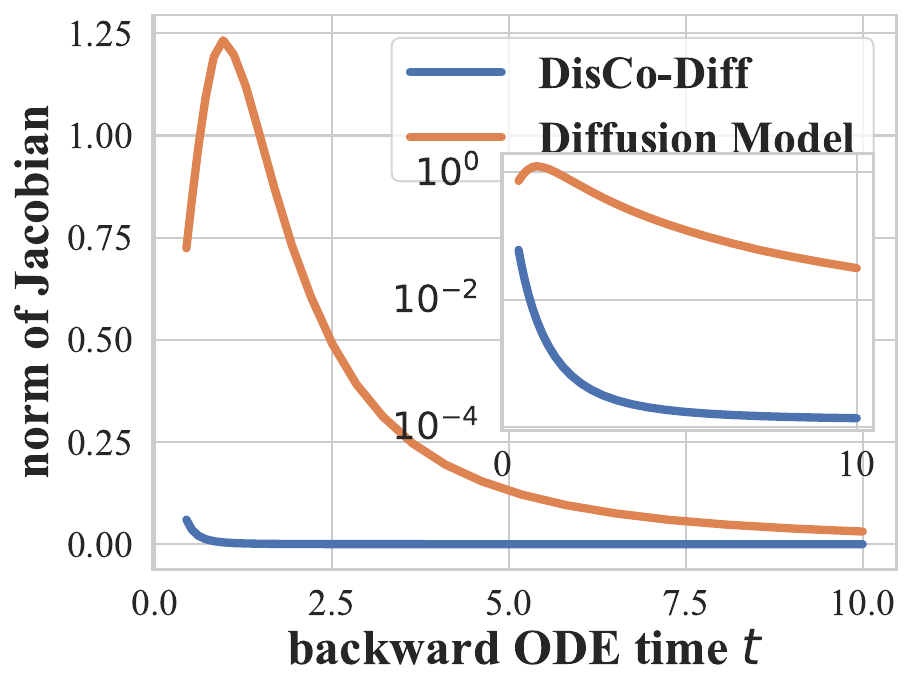}}
        \vspace{-5mm}
    \caption{\small \textbf{Modeling 2D mixture of Gaussians: analysis.} The mean curvature \textit{(left)} and norm of the neural networks' Jacobians \textit{(right)} along the reverse-time ODE trajectories as function of $t$.}
    \label{fig:toy-analysis}
    \vspace{-2mm}
\end{figure}
\textbf{A simpler mapping with discrete latent variables.} 
The role of the discrete latents in DisCo-Diff is to reduce this complexity and make the DM's learning task easier. The single noise-to-data mapping is effectively partitioned into a set of simpler mappings, each with less curvature in its generative ODE. We argue that it is unnatural to map an entire multimodal complex data distribution to a single continuous Gaussian distribution. Instead, we believe that an ideal generative model should combine both discrete and continuous latent variables, where discrete latents capture global multimodal structure and the continuous latents model local continuous variability. With this in mind, we suggest to only use a moderate number of discrete latents with small codebooks. On the one hand, a few latents can already significantly simplify the DM's learning task. On the other hand, if we only have few latents with small codebooks, training a generative model---an autoregressive one in our case---over the discrete latent variable distribution itself, will be simple (which we observe, see \Cref{sec:experiments}).  

\textbf{Validation in 2D.} To validate our reasoning, let us revisit the toy 2D mixture of Gaussians. 
In \Cref{fig:toy-vis}, \textit{right}, we show the DisCo-Diff model's synthesized data. The discrete latents learn to capture the different modes, and DisCo-Diff's DM component models the individual modes.  
The DM's ODE trajectories for different latents are now almost perfectly straight, indicating a simple conditional score function. In \Cref{fig:toy-analysis}, \textit{left}, we quantitatively show strongly reduced curvature along the entire diffusion time $t$. In \Cref{fig:toy-analysis}, \textit{right}, as a measure of network complexity we also show the norms of the Jacobians of the employed denoiser networks. We see significantly reduced norms for DisCo-Diff for all $t$, suggesting that the denoiser's task is indeed strongly simplified and less network capacity is required.\looseness=-1

\begin{figure}[t!]
  \begin{center}
    \includegraphics[width=0.85\columnwidth]{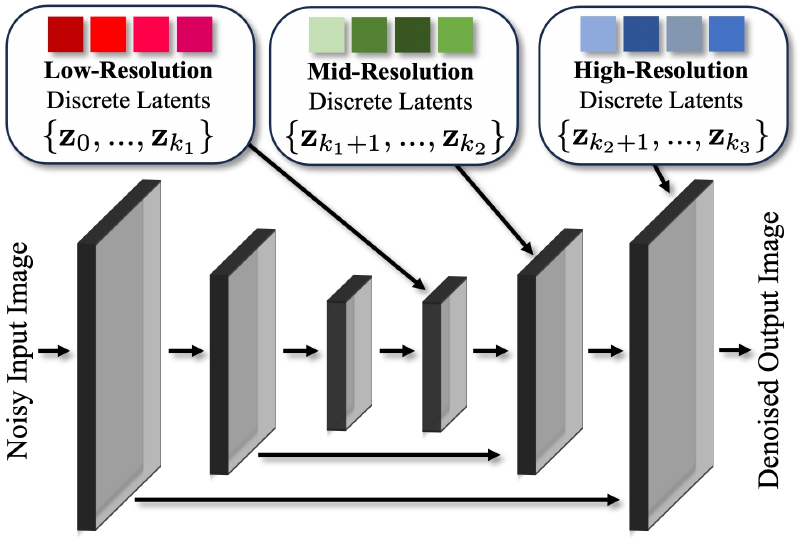}
  \end{center}
  \vspace{-5mm}
    \caption{\small \textbf{Group hierarchical DisCo-Diff.} Different discrete latents are fed to the denoiser U-Net at different feature resolutions.}
    \label{fig:hierarchical_arch}
  \vspace{-3mm}
\end{figure}
\label{sec:disco_arch}
\textbf{Using few, global latents with relatively small codebooks is important.}
DisCo-Diff is fundamentally different from most contemporary generative modeling frameworks using discrete latent variables~\cite{vandenoord2017vqvae,esser2021vqgan,ramesh2021dalle,chang2022maskgit,yu2022scaling,pernias2023wuerstchen,chang2023muse}. These works use autoencoders to encode images in its entirety into spatially-arranged, downsampled representations of the inputs, focusing more on preserving image fidelity than on capturing diverse training data modes. 
However, this is also unnatural: Encoding continuous variability, like smooth pose, shape, or color variations in images, into discrete latents requires the use of very large codebooks and, on top of that, these models generally rely on very high-dimensional spatial grids of discrete latents (\textit{e.g.} 32x32=1024 latents with codebooks ${>}1,000$~\cite{esser2021vqgan}, while we use just 10 latents with a codebook size of $100$ in our main image models). This makes learning the distribution over the discrete latents very challenging for these types of models, while it is simple in DisCo-Diff, where they just supplement the DM. 
In DisCo-Diff, we get the best from both continuous and discrete latent variables, using only few \textit{global} latents.\looseness=-1

\textbf{End-to-end training is essential.} DisCo-Diff's discrete latents are in spirit similar to leveraging non-learnt conditioning information. As pointed out by \citet{bao2022conditional}, this has been crucial to facilitate training high-performance generative models like strong class-conditional~\cite{dhariwal2021diffusion,Kingma2023UnderstandingDO} or text-to-image DMs~\cite{ramesh2022dalle2,ho2022imagen,rombach2021highresolution}. However, DisCo-Diff aims to fundamentally address the problem, rather than relying on given conditioning data. Moreover, the data usually has significant variability even given, for instance, a class label. Our discrete latents can further reduce the complexity (as observed, see \Cref{sec:experiments}).

However, could we use pre-trained encoder networks, such as CLIP~\cite{radford2021clip} or others~\cite{he2020moco,caron2021dino}, to produce encodings to condition on and whose distribution could be modeled in a second stage? This is explored by previous works~\cite{harvey2023visual,bao2022conditional,hu2023SelfGuidedDM, li2023selfconditioned}, but has important disadvantages: \textit{(i)} The most crucial downside is that such encoders are not universally available, but typically only for images. However, we seek to develop a universally applicable framework. For instance, we also apply DisCo-Diff to molecular docking (see \Cref{sec:experiments_docking}), where no suitable pre-trained networks are available. \textit{(ii)} In DisCo-Diff, the job of the discrete latents is to make the denoising task of the DM easier, which is especially ambiguous at large noise levels (in fact, we find that the latents help in particular to reduce the loss at these high noise levels, see \Cref{fig:loss_analysis}). It is not obvious what information about the data the latents should best encode for this. By learning them jointly with the DM objective itself, they are directly trained to help the DM learn better denoisers and lower curvature generative ODEs. \textit{(iii)} A generative model needs to be trained over the encodings in the second stage. In DisCo-Diff, we can freely choose an appropriate number of latents and codebook size to simplify the DM's denoising task, while also facilitating easy learning of the autoregressive model in the second stage. When using pre-trained encoders, one must work with the encodings by these methods, which were not developed for generative modeling. 
We attribute DisCo-Diff's strong generation performance to its end-to-end learning.

\textbf{The latent variables must be discrete.} Could we also use auxiliary continuous latent variables? Generative models on continuous latents are almost always based on mappings from a uni-modal Gaussian distribution to the distribution of latents. Hence, if such continuous latents learnt multimodal structure in the data to simplify the main DM's denoising task, as DisCo-Diff's discrete latents do, then learning a distribution over them in the second stage would again require a highly non-linear difficult-to-learn mapping from Gaussian noise to the multimodal encodings. This is the problem DisCo-Diff aims to solve in the first place. \citet{preechakul2021diffusion} augment DMs with non-spatial continuous latent variables, but they only focus on semantic face image manipulation.
InfoDiffusion~\cite{wang2023infodiffusion} conditions DMs on discrete latent variables. However, it focuses on learning disentangled representations, also primarily for low-resolution face synthesis, and uses a mutual information-based objective. Contrary to DisCo-Diff, neither of these works tackles high-quality synthesis for challenging, diverse datasets. 

In our ablation studies (\Cref{sec:image_synthesis}), we further validate our design choices and motivations that we presented here.
\begin{table}[t!]
    \vspace{-3mm}
    \small
    \centering
    \caption{FID score together with NFE on ImageNet-64.}
    \scalebox{0.9}{\begin{tabular}{l c c}
    \toprule
    & FID & NFE \\
    \midrule\midrule
    \textit{without class-conditioning}\\
    \midrule
    IC-GAN~\cite{Casanova2021InstanceConditionedG} & 9.20 & 1 \\
    BigGAN~\cite{Brock2018LargeSG} & 16.90 & 1 \\
    iDDPM~\cite{Nichol2021ImprovedDD} & 16.38 & 50 \\
    EDM~\cite{Karras2022ElucidatingTD} & 6.20 & 50 \\
    SCDM~\cite{bao2022conditional} & 3.94 & 50\\
    DisCo-Diff (\textit{ours})& \textbf{3.70} & 50\\
    \midrule\midrule
    \textit{class-conditioned, ODE sampler}\\
    \midrule
    EDM~\cite{Karras2022ElucidatingTD} & 2.36 & 79\\
    PFGM++~\cite{xu2023pfgmpp} & 2.32 & 79\\
    DisCo-PFGM++ (\textit{ours}) & 1.92 & 78\\
     DisCo-Diff (\textit{ours}) & \textbf{1.65} & 78\\
    \midrule
    \textit{class-conditioned, stochastic sampler}\\
    \midrule
    iDDPM~\cite{Nichol2021ImprovedDD} & 2.92 & 250 \\
ADM~\cite{dhariwal2021diffusion} & 2.07 & 250 \\
CDM~\cite{Ho2021CascadedDM} & 1.48 & 8000\\
VDM++~\cite{Kingma2023UnderstandingDO} & 1.43 & 511\\
EDM (w/ Restart~\cite{Xu2023RestartSF}) & 1.36 & 623 \\
RIN~\cite{Jabri2022ScalableAC} & 1.23 & 1000\\
 DisCo-Diff  (\textit{ours}; w/ Restart~\cite{Xu2023RestartSF}) & \textbf{1.22} & 623 \\
\midrule
    \textit{class-conditioned, w/ adversarial objective}\\
    \midrule
    IC-GAN~\cite{Casanova2021InstanceConditionedG} & 6.70 & 1 \\
    BigGAN-deep~\cite{Brock2018LargeSG} & 4.06 & 1\\
    CTM~\cite{Kim2023ConsistencyTM} & 1.92 & 1\\
    StyleGAN-XL~\cite{Sauer2022StyleGANXLSS} & 1.51 & 1\\
    \bottomrule
    \end{tabular}}
    \label{tab:imagenet-64}
    \vspace{-3mm}
\end{table}
\vspace{-1mm}
\subsection{Architecture}\label{sec:disco_arch}
\vspace{-1mm}



As discussed,
DisCo-Diff enhances the training of continuous DMs by incorporating learnable discrete latent variables that are meant to capture the \textit{global} underlying discrete structure of the data. 
To ensure that DisCo-Diff works as intended, suitable network architectures are necessary. Below, we summarize our design choices, focusing on DisCo-Diff for image synthesis. However, the framework is general, requiring only an encoder to infer discrete latents from clean input data and a conditioning mechanism that integrates these discrete latents into the denoiser network. In fact, we also apply our model to 2D toy data and molecular docking.


\textbf{Encoder.} For image modeling, we utilize a ViT~\cite{dosovitskiy2021vit} as the backbone for the encoder. We extend the classification mechanism in ViTs, and treat each discrete token as a different classification token. Concretely, we add $m$ extra classification tokens to the sequence of image patches. This architectural design naturally allows each discrete latent to effectively capture the global characteristic of the images, akin to performing data classification. 


\textbf{Discrete latent variable conditioning.} For image experiments, DisCo-Diff's denoisers are U-Nets as widely used for DMs~\cite{Karras2022ElucidatingTD,hoogeboom2023simplediffusion}. For the discrete latent variable conditioning, we utilize cross-attention~\cite{rombach2021highresolution}.
Drawing inspiration from text-to-image generation, DisCo-Diff's discrete latents function analogously to text, exerting a global influence on the denoiser's output. Specifically, image features act as queries and discrete latents are keys and values in the cross-attention layer, enabling discrete latents to globally shape the image features. 
We add a cross-attention layer after each self-attention layer within the U-Net. 
In our main models, all discrete latents are given to all cross-attention layers.

\textbf{Group hierarchical models.} To enhance the interpretability of discrete latents, we also explore the inductive bias inherent in the U-Net architecture and feed distinct latent groups into various resolution features in the up-sampling branch of the U-Net, as shown in \Cref{fig:hierarchical_arch}. This approach draws inspiration from StyleGAN~\cite{karras2019stylegan}, where distinct latents are introduced at different resolutions, enabling each to capture different image characteristics by the neural network's inductive bias. This design fosters a group hierarchy, where the groups associated with higher-resolution features offer supplementary information, conditioned upon the groups related to lower-resolution features. We refer to this refined model as the \textit{group hierarchical} DisCo-Diff.


\begin{table}[t]
  \vspace{-3mm}
    \small
    \centering
    \caption{FID score and NFE on class-cond. ImageNet-128.}
    \scalebox{0.89}{\begin{tabular}{l c c}
    \toprule
    & FID & NFE \\
    \midrule
    ADM~\cite{dhariwal2021diffusion}
    & 5.91 & 250 \\   
    ADM-G~\cite{dhariwal2021diffusion}
    & 2.97 & 250 \\ 
    CDM (32, 64, 128)~\cite{Ho2021CascadedDM} & 3.52 & 8100\\
    RIN~\cite{Jabri2022ScalableAC} & 2.75 & 1000 \\
    \midrule
    VDM++, \textit{w/ ODE sampler}~\cite{Kingma2023UnderstandingDO} & 2.29 & 115\\
    DisCo-Diff, \textit{w/ ODE sampler} (\textit{ours}) & {1.98} & 114\\
    VDM++, \textit{w/ DDPM sampler}~\cite{Kingma2023UnderstandingDO} & 1.88 & 512\\
    DisCo-Diff, \textit{w/ ODE sampler, VDM++ correction} & \textbf{1.73} & 414\\
    \bottomrule
    \end{tabular}}
    \label{tab:imagenet-128}
    \vspace{-4mm}
\end{table}
In the \textbf{molecular docking} task, existing denoisers operate through message passing in a permutation equivariant way over 3D point clouds representing molecular structures~\cite{corso2023diffdock}. We build this property and architectural bias directly into the latent variables, allowing them to take values indicating one node in the point cloud (therefore, for every point cloud, the codebook size equals the number of nodes). This latent design choice aligns with the intuition of the encoder determining the atoms playing key roles in the structure and allows for minimal modification of the score model where the latents simply represent additional features for every node. The encoder is also composed of a similar equivariant message passing, \texttt{e3nn}~\cite{geiger2022e3nn}, network where for each node one logit per latent will be predicted. More details on the architecture for the molecular docking task can be found in \Cref{app:docking_architecture}. 

The \textbf{auto-regressive model} over the distribution of the discrete latents is implemented in image experiments using a standard Transformer decoder~\cite{Vaswani2017AttentionIA}. For molecular docking, it again uses an \texttt{e3nn} network that is fed the conditioning information of the protein structure and molecular graph. 
Generally, DisCo-Diff is compatible with other conditional inputs, \textit{e.g.} class labels, which can be added as inputs to denoiser and auto-regressive model. We use an auto-regressive model for simplicity and expect DisCo-Diff's second stage to work equally well with other discrete data generative models, \textit{e.g.} discrete state diffusion models~\cite{austin2021structured,campbell2022a}. 
Architecture details, also for 2D toy data experiments, in \Cref{app:exps_details}.\looseness=-1

An important question surrounding the architecture design is \textbf{how to choose an appropriate number of latents and codebook size.} While intuitively, increasing the number of latents and the codebook size might seem like a straightforward method to reduce the reconstruction error further by capturing more intricate data structures, this approach also introduces additional complexity. Specifically, a larger set of latents or an expanded codebook size complicates the auto-regressive model's task, potentially leading to increased errors. Thus, finding a balance between enhancing model performance through more detailed discrete structures and maintaining manageable modeling complexity for the auto-regressive model is crucial. We recommend using a modest number of latent (e.g. 10~30) and codebook size (e.g. 50~100) for current diffusion models and leaving the exploration of optimal hyper-parameters to future works. For example, in our image generation experiments, we found that a configuration of 10 latents with a codebook size of 100 significantly enhances performance on the complex ImageNet dataset. We did not tune this hyper-parameter due to computational constraints. We believe that more careful hyper-parameter optimization over the exact number of latents and the codebook size would further boost the performance of DisCo-Diff.

\begin{table}[t]
    \small
    \centering
    \caption{Ablations on class-cond. ImageNet-64.}
    \scalebox{0.9}{\begin{tabular}{l c c}
    \toprule
    & FID  \\
    \midrule
    EDM~\cite{Karras2022ElucidatingTD} & 2.36 \\
    \midrule
    \textit{Oracle setting} \\
    \midrule
    Continuous latent (KLD weight=0.1) & 1.67  \\
    Continuous latent (KLD weight=1) & 2.36  \\
    DisCo-Diff (cfg=0) & 1.65\\
    \midrule
    \textit{Generative prior on latent} \\
    \midrule
    Continuous latent (KLD weight=0.1) & 11.12  \\
    Continuous latent (KLD weight=1, cfg=0) & 2.36  \\
    Continuous latent (KLD weight=1, cfg=1) &  2.36 \\
    DisCo-Diff (cfg=0) & 1.81 \\
    DisCo-Diff (cfg=1) & \textbf{1.65} \\
    DisCo-Diff (cfg=2) & 2.33 \\
    \bottomrule
    \end{tabular}}
    \label{tab:imagenet-64-ablation}
    \vspace{-2mm}
\end{table}

\section{Experiments}\label{sec:experiments}

\subsection{Image Synthesis} \label{sec:image_synthesis}
We use the ImageNet~\cite{deng2009imagenet} dataset and tackle both class-conditional (at varying resolutions~$64{\times} 64$ and $128{\times} 128$) and unconditional synthesis. To measure sample quality, we follow the literature and use Fr\'{e}chet Inception Distance (FID)~\cite{heusel2017gans}~(lower is better). We also report the number of neural function evaluations (NFE).

In the class-conditional setting, the DisCo-Diff's denoiser is initialized using pre-trained ImageNet models, except for the new components: the cross-attention layers between discrete latents and images, and the encoder. We fine-tune the pre-trained U-Net in EDM~\cite{Karras2022ElucidatingTD} with discrete latents for ImageNet-64. For ImageNet-128, we implement the U-ViT in VDM++~\cite{Kingma2023UnderstandingDO}, and fine-tune our trained VDM++ model using discrete latents.
We also adhere to their respective noise schedules and loss weightings during the training process. We use Heun's second-order method as  ODE sampler, and a 12-layer Transformer as the auto-regressive model. We set the latent dimension to $m=10$ and the codebook size to $k=100$ in DisCo-Diff.
\textbf{Results.} See \Cref{tab:imagenet-64,tab:imagenet-128}.
\textbf{(1) DisCo-Diff achieves the new state-of-the-art on class-conditioned ImageNet-64/ImageNet-128 when using ODE sampler.} Specifically, DisCo-Diff reduces the previous state-of-the-art FID score from 2.36 to 1.65 on ImageNet-64, and from 2.29 to 1.98 on ImageNet-128. This aligns with our analysis (\Cref{sec:disco_motivation}) 
that DisCo-Diff yields straighter ODE trajectories.

\textbf{(2)~DisCo-Diff outperforms all baselines in the unconditional setting, or when using stochastic sampler.} DisCo-Diff also surpasses the previous best method~(SCDM~\cite{bao2022conditional}) in the unconditional setting, even though their method relies on pre-trained MoCo features. In addition, DisCo-Diff sets the new record ImageNet-64 FID of 1.22 when using Restart sampler~\cite{Xu2023RestartSF}. Note that the competitive method RIN~\cite{Jabri2022ScalableAC} employs a novel architecture distinct from conventional U-Nets/U-ViTs. 

\begin{figure*}[t]
    \centering
    \subfigure{\includegraphics[width=0.24\textwidth]{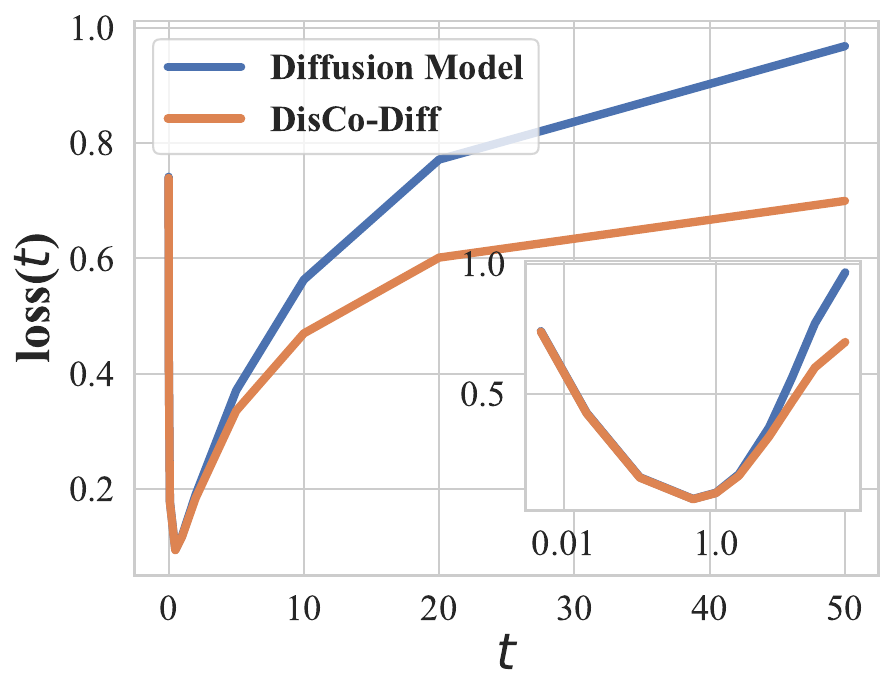}}\hspace{1mm}    \subfigure{\includegraphics[width=0.64\textwidth]{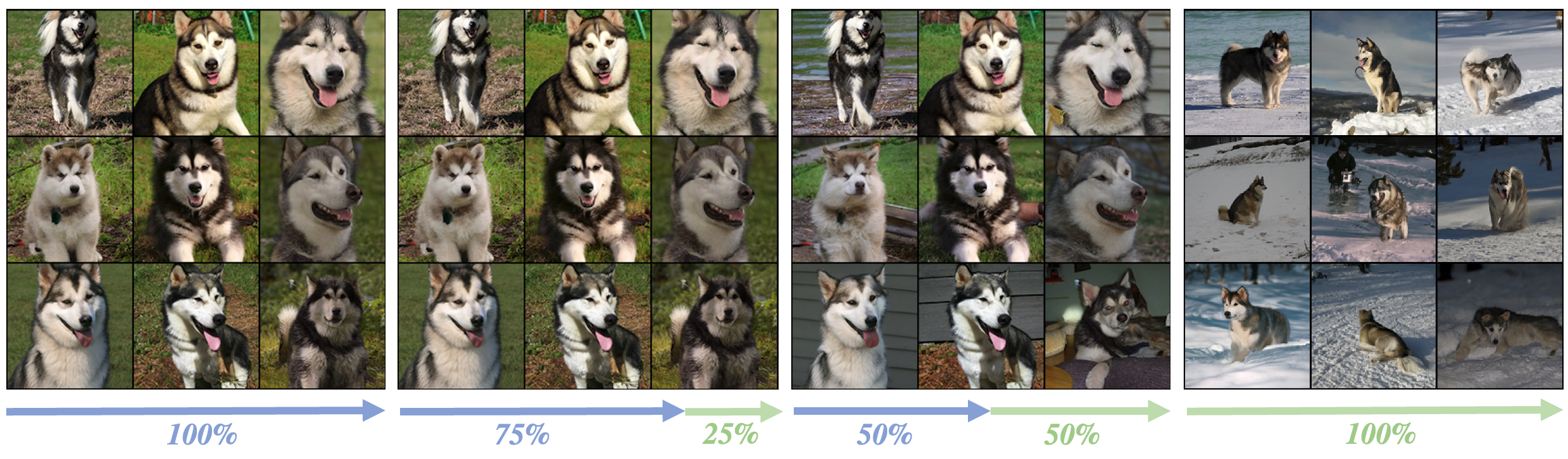}}
    \vspace{-5mm}
    \caption{\textit{Left:} Loss versus time. \textit{Right:} Impact of discrete latent switching during the iterative sampling process of DisCo-Diff's diffusion model component. The numbers represent the percentage of the total sampling steps. {The blue/green arrows mean the sampling steps that utilize the discrete latent associated with the leftmost/rightmost grid in the figure. For the middle two images, the process involves initially employing the discrete latent from the leftmost grid for a certain proportion of the total sampling steps (e.g., 75\%), before transitioning to the discrete latent from the rightmost grid to complete the remaining steps (e.g., the last 25\% of the total sampling steps).}}
    \label{fig:loss_analysis}
    \vspace{-5pt}
\end{figure*}

{
On ImageNet-128, we observe that DisCo-Diff does not perform well with stochastic samplers. When using the DDPM sampler as in VDM++~\cite{Kingma2023UnderstandingDO}, DisCo-Diff achieves an FID score of 2.80, which is worse than VDM++'s FID score of 1.88. As the discrete latents reduce the loss at larger times~(\textit{c.f.} \Figref{fig:loss_analysis}) and the training targets at these times typically correspond to low-frequency components of images, we hypothesize that in this model the discrete latents learnt to overly emphasize global information, diverting the model to overlook some high-frequency details necessary at smaller times. We provide empirical evidence in Appendix~\ref{app:imagenet-128} to show that VDM++ w/ DDPM better captures details at smaller times, which supports this hypothesis. Note that, in theory, VDM++ and DisCo-Diff should perform similarly at smaller times. A potential solution could concatenate the discrete latents with a null token, similar to text-to-image models~\cite{balaji2022ediffi}, allowing the model to learn more easily to exclude the influence of discrete latents at smaller times. We leave it for future exploration. We would like to emphasize that we only observed this behavior for this one model and dataset. In all other experiments, discrete latents universally improved performance for all stochastic and non-stochastic samplers, and when used for all times $t$.
To better utilize the DDPM sampler for the current model, we substituted the DisCo-Diff ODE with the VDM++ DDPM trajectories at smaller times~($t<10$), (DisCo-Diff, \textit{w/ ODE sampler, VDM++ correction} in Table~\ref{tab:imagenet-128}), which improves the FID to 1.73. In conclusion, while the discrete latents did not help at small times here, they still boosted performance at larger times and allowed us to outperform the pure VDM++ model and, once again, achieve state-of-the-art performance.} 

\textbf{(3)~Discrete latents capture variability complementary to class semantics.} \Cref{fig:vis-128} (b) illustrates that samples sharing the same discrete latent exhibit similar characteristics, and there are noticeable distinctions for different discrete latents under the same class. It suggests that the discrete latents capture variations that are useful in simplifying the diffusion process defined in Euclidean space beyond class labels, underpinning the improvements of DisCo-Diff over the pre-trained class-conditioned DMs. 
\textbf{(4)~Discrete latents boost the performance on PFGM++}. When applied to another ODE-based generative model PFGM++~\cite{xu2023pfgmpp}, DisCo-PFGM++ also improves over the baseline version (see \Cref{tab:imagenet-64}).
More samples in \Cref{app:extra_sample}.

\begin{figure}[t!]
    \centering
    \includegraphics[width=0.45\textwidth]{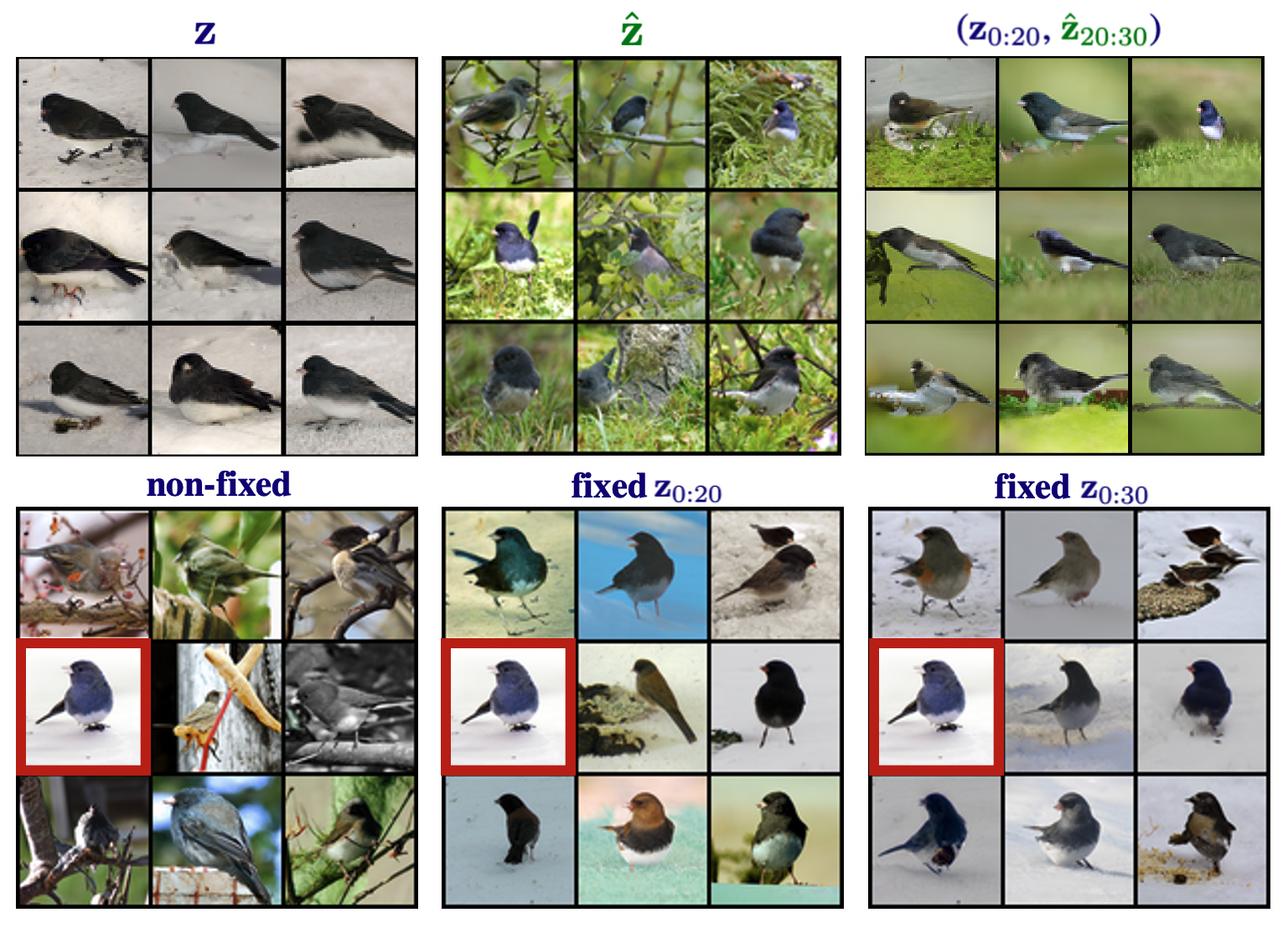}
     \vspace{-4mm}
    \caption{\small \textit{Top:} Images created from two $30$-dim discrete latents {\color{black!50!blue}{$\rvz$}} and {\color{black!50!green}{$\hat{\rvz}$}}, with the far-right column combining their sub-coordinates. \textit{Bottom:} Variations in images by fixing portions of {\color{black!50!blue}{$\rvz$}} (originating from the red-boxed image). We see that lower-resolution latents affect layout / shape; high-resolution latents alter color / texture. }
    \label{fig:vis-hierarchy}
    \vspace{-2mm}
\end{figure}

\begin{figure}[b!]
\vspace{-3mm}
    \centering
    \includegraphics[width=0.48\textwidth]{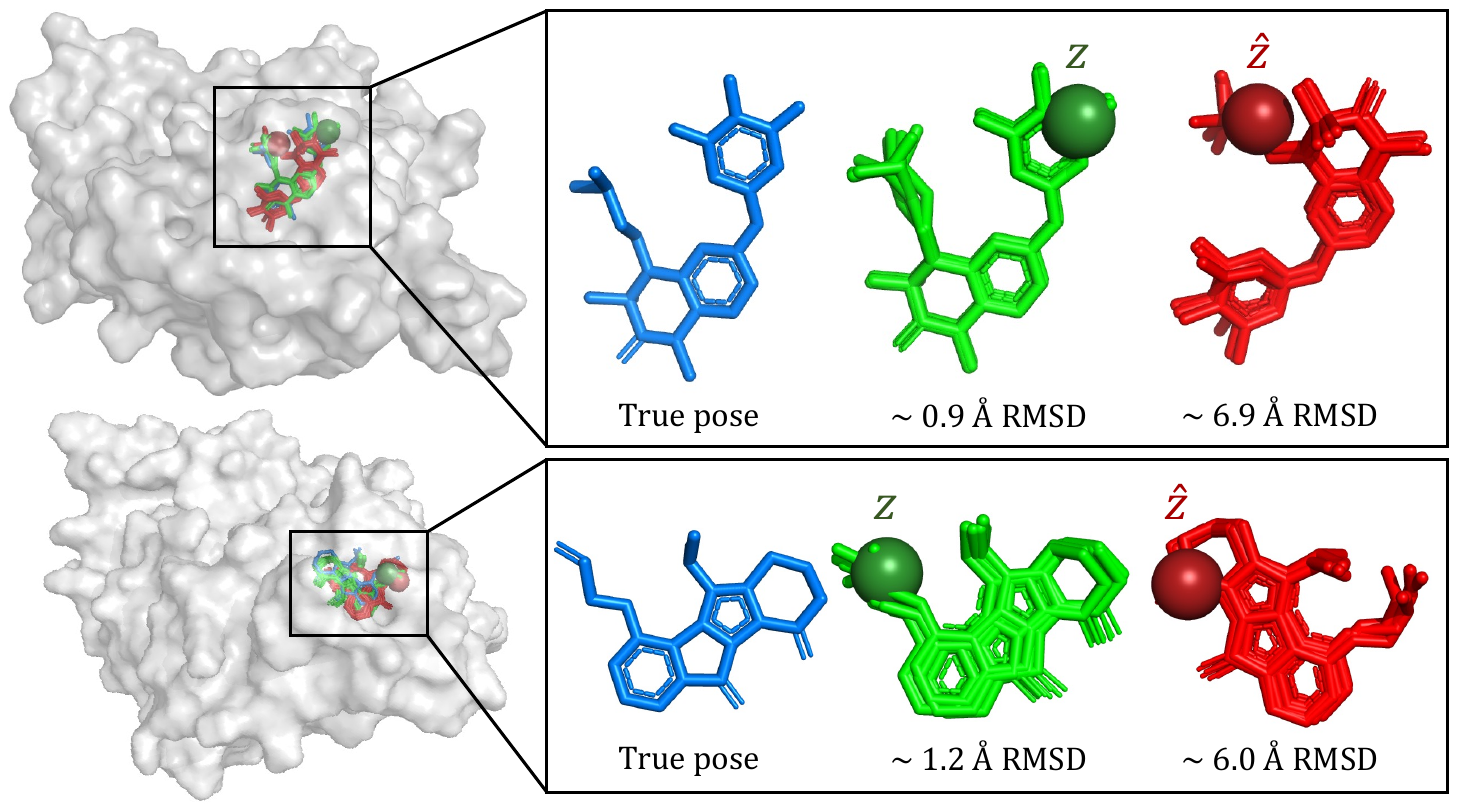}
     \vspace{-7mm}
    \caption{\small  Examples of alternative docking poses modeled when conditioning on different discrete latents, the ``correct" $z$ (i.e. same as the encoder) and an incorrect $\hat{z}$. The DM maps them to two distinct sets of orientations with which the ligand could fit in the pocket. Notably, the correct latent corresponds to poses within 2\AA{} of the ground truth. 
    The colored beads are set on the atoms corresponding to the first latent variable.}
    \label{fig:vis-docking}
    \vspace{-1mm}
\end{figure}
\textbf{Ablations and Analyses.}
Table~\ref{tab:imagenet-64-ablation} shows that employing moderate classifier-free guidance with respect to the discrete latents (scale cfg=1) enhances the FID score (studied using ODE sampler), implying
that the discrete latents effectively learn modes similar to the role of class labels and text.
We further substituted the discrete latents with 1000-dim. continuous latents (1000 to offer capacity at least as high as with the $m{=}10$ and $k{=}100$ discrete latents), using Kullback-Leibler divergence-based (KLD) regularization as in VAEs to control the information retained.
For fair comparison, we trained a DiT-based DM~\cite{peebles2023scalable} on the continuous latents using the same Transformer architecture as in DisCo-Diff's auto-regressive model.
\Cref{tab:imagenet-64-ablation} shows that with a low KLD weight (0.1), the continuous latents are under-regularized, challenging the DiT in modeling the complex encoding distribution and leading to a significant gap between oracle FID~(latents predicted from training images) and generative FID~(latents sampled from second-stage latent generative models). Conversely, a higher KLD weight~(1) causes encoder collapse, and the continuous latents are not used (no latent (EDM), oracle latents and generative latents all produce same FIDs). In contrast, DisCo-Diff's generative FID shows only a minor degradation compared to the oracle FID, indicating the ease of modeling the discrete prior with a simple Transformer. 

The DM training objective (\Cref{eq:dsm}) has most variability at large diffusion times due to the multimodal posterior of clean data given noisy inputs~\cite{Xu2023StableTF}. Conditioning information can 
reduce this ambiguity. For instance, \citet{balaji2022ediffi} show that text conditioning primarily influences the denoiser at larger times.
\Cref{fig:loss_analysis} (a) shows that the learned discrete latents behave similarly to text conditioning, significantly lowering the training loss at higher time steps. Complementarily, \Cref{fig:loss_analysis} (b) indicates that switching discrete latents towards the end of sampling barely affects the samples, implying they are not used at smaller times $t$.

In DisCo-Diff, the sampling time of the auto-regressive model is negligible compared to the DM's. For instance, for generating 32 images on ImageNet-128, the auto-regressive models requires only 0.44 seconds, while DisCo-Diff's DM component takes 78 seconds for 114 NFE, with an average of 0.68 second/NFE, all on a single NVIDIA A100 GPU.


\textbf{Group Hierarchical DisCo-Diff.} We evaluate the group hierarchical DisCo-Diff (\Cref{sec:disco_arch}),
feeding three separate 10-dim. discrete latents into the U-Net at each level of resolution. Fig.~\ref{fig:vis-hierarchy} shows that latents for lower-resolution features mainly govern overall shape and layout, while latents for higher-resolution control color and texture. For example, in the bottom figure, when gradually fixing groups in order, the images first converge in shape and then in color. 

\vspace{-1mm}
\subsection{Molecular Docking}\label{sec:experiments_docking}
\vspace{-1mm}
We test DisCo-Diff also on molecular docking, a fundamental task in drug discovery that consists of generating the 3D structure with which a small molecule will bind to a protein. We build on top of DiffDock \cite{corso2023diffdock}, a DM that recently achieved state-of-the-art performance, integrating discrete latent variables (see Sec.~\ref{sec:method} and App.~\ref{app:docking_architecture} for details).
For computational reasons, we use the reduced DiffDock's architecture (referred to as DiffDock-S) from \citet{corso2024discovery}, which, although less accurate, is much faster for training and inference. For training and evaluation, we follow the standard from \citet{equibind} using the PDBBind dataset \cite{liu2017PDBBind} (see App.~\ref{app:docking_exps_details}).



\textbf{Results.} \Cref{tab:docking_results} reports performance of our (\textit{DisCo-DiffDock-S}) and relevant baseline methods.
We see that also in this domain discrete latents provide improvements, 
with the success rate on the full dataset increasing from 32.9\% to 35.4\% and from 13.9\% to 18.5\% when considering only test complexes with unseen proteins.
This improvement is particularly strong on the harder component of the test set, where the baseline model is, likely, highly uncertain. 
This supports the intuition that DisCo-Diff boosts performance by more appropriately modeling discrete and continuous variations in the data.
In Fig.~\ref{fig:vis-docking}, we visualize two examples from the test set which highlight how the model learns to associate distinct sets of poses with different latents, decomposing the multimodal components of the pose distribution from the continuous variations that each pose can have.

\begin{table}[h]
\vspace{-4mm}
    \small
    \centering
    \caption{Molecular docking performance on PDBBind. For each method, we report the percentage of top-1 predictions within 2\AA{} of the ground truth for the \textit{full} test set and the subset restricted to \textit{unseen} proteins. Runtime in seconds (* refers to run on CPU). } \label{tab:docking_results}

\scalebox{0.9}{\begin{tabular}{lccc} \toprule
           & Full & Unseen & Runtime    \\ \midrule
GNINA \citep{mcnutt2021gnina}           & 22.9          & 14.0           & 127  \\
SMINA \citep{koes2013smina}           & 18.7          & 14.0           & 126*       \\
GLIDE \citep{halgren2004glide}           & 21.8          & 19.6           & 1405*      \\
EquiBind \citep{equibind}        & 5.5           & 0.7            & 0.04 \\
TankBind \citep{Lu2022TankBind}        & 20.4          & 6.3            & 0.7  \\ \midrule
DiffDock-S \citep{corso2024discovery}      & 32.9          & 13.9           &   8.1         \\
DisCo-DiffDock-S (\textit{ours})   & 35.4          & 18.5           &   9.1          \\   \midrule    
DiffDock \citep{corso2023diffdock}    & 38.2          & 20.8           & 40   \\  \bottomrule
\end{tabular}}
\vspace{-3mm}
\end{table}

\section{Conclusions}

We have proposed \textit{\textbf{Dis}crete-\textbf{Co}ntinuous Latent Variable \textbf{Diff}usion Models (DisCo-Diff)}, a novel and universal framework for combining discrete latent variables with continuous DMs. The approach significantly boosts performance by simplifying the DM's denoising task through the help of auxiliary discrete latent variables, while introducing negligible overhead. Extensive experiments and analyses demonstrate the unique benefits of global discrete latent variables that are learnt end-to-end with the denoiser. DisCo-Diff does not rely on any pre-trained encoder networks. As such, we validated our method not only on image synthesis, but also for molecular docking, demonstrating its universality. 

\textbf{Limitations and Future Work.} There are several potential future directions and limitations in both the experiments and design of DisCo-Diff. First, our experiments have been primarily focused on standard benchmarks such as ImageNet. With more compute resources, DisCo-Diff could be further validated on tasks such as text-to-image generation, where we would expect discrete latent variables to offer complementary benefits to the text conditioning, similar to how discrete latents boost performance in our class-conditional experiments. Secondly, the Group Hierarchical model relies on inductive biases in its architecture, such as the different image characteristics captured at different resolutions in the U-Net. It would be interesting to explore how such architectures could be constructed and similar hierarchical effects could be achieved when working with different data modalities (molecules, etc.). Thirdly, one could apply the idea of DisCo-Diff to other continuous flow models, such as flow-matching~\cite{Lipman2022FlowMF} or rectified flow~\cite{liu2022flow}, to further boost their performance. Conceptually, due to the close relation between diffusion models and flow matching, we expect discrete latents to behave similarly there and improve performance. Finally, the current DisCo-Diff framework leverages a two-stage training process. Initially, we jointly train the denoiser and the encoder, followed by the post-hoc auto-regressive model in the second stage. Future work could investigate combining the two-stage training into a seamless end-to-end fashion.


\clearpage

\section*{Impact Statement}
Deep generative modeling is a burgeoning research field with widespread implications for science and society. Our model DisCo-Diff advances the modeling power of diffusion models for data generation. While enhancing data generation capabilities, notably in high-quality image and video creation, these models also present challenges, such as the potential misuse in deepfake technology leading to social engineering concerns. Addressing these issues necessitates further research into watermark algorithms for diffusion models and collaboration with socio-technical disciplines to balance innovation with ethical considerations. We would also like to highlight the promise of deep generative models like DisCo-Diff in the natural sciences, as exemplified by our molecular docking experiments. Such models have the potential to provide novel insights into, for instance, the interactions between proteins and ligands and advance drug discovery.\looseness=-1

\section*{Acknowledgements}

YX and TJ acknowledge support from MIT-DSTA Singapore collaboration; YX, GC, and TJ from NSF Expeditions grant (award 1918839) “Understanding the World Through Code”, from MIT-IBM Grand Challenge project, NSF Expeditions grant (award 1918839: Collaborative Research: Understanding the World Through Code), Machine Learning for Pharmaceutical Discovery and Synthesis (MLPDS) consortium; GC and TJ further acknowledge support from the Abdul Latif Jameel Clinic for Machine Learning in Health and the DTRA Discovery of Medical Countermeasures Against New and Emerging (DOMANE) threats program.\looseness=-1

\bibliography{bibliography}
\bibliographystyle{icml2024}

\newpage
\appendix
\onecolumn

\renewcommand\ptctitle{}
\addcontentsline{toc}{section}{Appendix} 
\part{Appendix} 

\parttoc 


\section{Related Work}\label{app:related_work}
Our work builds on DMs~\cite{sohl2015deep,ho2020ddpm,song2020score,Karras2022ElucidatingTD}, which have been widely used not only for image generation~\cite{dhariwal2021diffusion,Nichol2021ImprovedDD,rombach2021highresolution,dockhorn2022score,dockhorn2022genie,saharia2022photorealistic,ramesh2022dalle2,podell2023sdxl}, but also for video~\cite{blattmann2023videoldm,singer2023makeavideo,ho2022imagen,ge2022pyoco}, 3D~\cite{nichol2022pointe,zeng2022lion,kim2023nfldm,poole2023dreamfusion,schwarz2023wildfusion,liu2023zero123} and 4D~\cite{singer2023mav3d,ling2024align,bahmani20234dfy,zheng2023unified} synthesis, as well as in various other domains, including, for instance, molecular docking and protein design~\cite{corso2023diffdock,yim2023framediff,ingraham2022chroma,watson2023rfdiffusion}.

In the DM literature, latent variables have been most popular as part of latent diffusion models, where a DM is trained in a compressed, usually continuous, latent space~\cite{rombach2021highresolution,vahdat2021score}. In contrast, DisCo-Diff leverages \textit{discrete} latent variables and uses them to augment a DM. The first models using discrete latent variables for high-dimensional generative modeling tasks include Boltzmann machines~\cite{salakhutdinov2009deep,hinton2012practical} and early discrete variational autoencoders~\cite{rolfe2017discrete,vahdat2018dvae,vahdat2018dvaepp}. More recently, a variety of works encode images into large 2D spatial grids of discrete tokens with vector quantization or similar techniques~\cite{vandenoord2017vqvae,esser2021vqgan,ramesh2021dalle,chang2022maskgit,yu2022scaling,pernias2023wuerstchen,chang2023muse}. As discussed, these models typically require a very large number of tokens and rely on large codebooks, which makes modeling their distribution challenging. DisCo-Diff, in contrast, leverages only few discrete latents with small codebooks that act in harmony with the additional continuous DM. 

There are previous related works that also condition DMs on auxiliary encodings. \citet{preechakul2021diffusion} augment DMs with non-spatial latent variables, but their latents are continuous and high-dimensional, which makes training their latent DM more challenging. This is precisely what we avoid by instead using low-dimensional and discrete latents. Moreover, they focus on semantic face image manipulation, not high-quality synthesis for challenging, diverse datasets. 

\citet{harvey2023visual} use the representations of a pre-trained CLIP image encoder~\cite{radford2021clip} for conditioning a DM and learn another DM over the CLIP embeddings for sampling. Similarly, \citet{bao2022conditional} and \citet{hu2023SelfGuidedDM} use clustered MoCo-based~\cite{he2020moco} and clustered DINO-based~\cite{caron2021dino} features, respectively, for conditioning. Hence, these three approaches are strictly limited to image synthesis, where such encoders, pre-trained on large-scale datasets, are available. In contrast, we purposefully avoid the use of pre-trained networks and learn the discrete latents jointly with the DM, making our framework universally applicable.
Another related work is InfoDiffusion~\cite{wang2023infodiffusion}, which also conditions DMs on discrete latent variables. However, contrary to DisCo-Diff, this work focuses on learning disentangled representations, similar to $\beta$-VAEs~\cite{higgins2017betavae}, primarily for low-resolution face synthesis. It uses a mutual information-based objective and does not focus on diverse and high-quality synthesis of complex data such as ImageNet. 

In contrast to the above works, we show how discrete latent variables boost generative performance itself and we significantly outperform these works in complex and diverse high-quality synthesis. Furthermore, we motivate DisCo-Diff fundamentally, with reduced ODE curvature and model complexity, providing a new and complementary perspective.\looseness=-1

In the molecular docking literature, since DiffDock \cite{corso2023diffdock} introduced the use of diffusion models in the task, a number of works have proposed different modifications to its framework. In particular, some \cite{mastersfusiondock, plainer2023diffdockpocket, guo2023diffdocksite} have proposed to separate the blind docking task between pocket identification (i.e. identifying the region of the protein where the small molecule would bind) and pose prediction (i.e. predicting the specific pose with which the ligand would bind to the protein), as previously done in many traditional approaches \cite{krivak2018p2rank}. One could see this as hand-crafting a (roughly discrete) latent variable in the pocket and using it to decompose the task. By allowing the encoder to learn arbitrary discrete latents through its interaction with the denoiser, DisCo-Diff largely includes the above-mentioned strategy as a particular case.

\section{Discrete Latent Variable Classifier-Free Guidance}\label{app:cfg}

\begin{figure}[H]
    \centering
    \subfigure[cfg=0]{\includegraphics[width=0.23\textwidth]{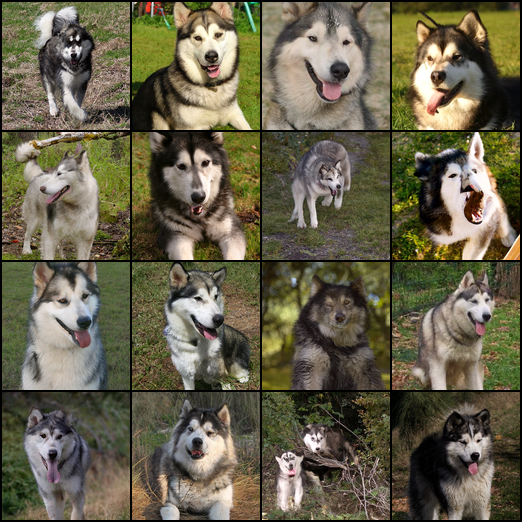}}\hfill
    \subfigure[cfg=1]{\includegraphics[width=0.23\textwidth]{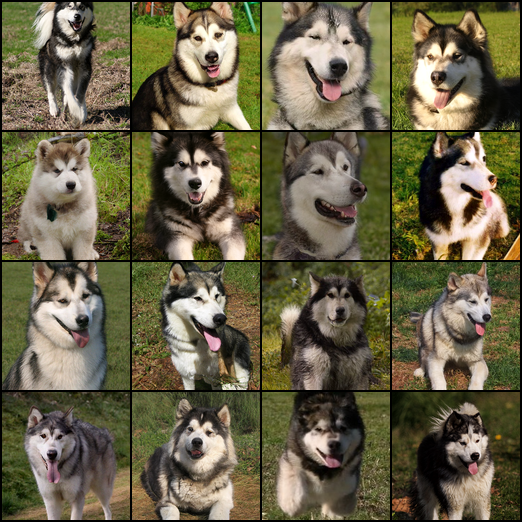}}\hfill
    \subfigure[cfg=4]{\includegraphics[width=0.23\textwidth]{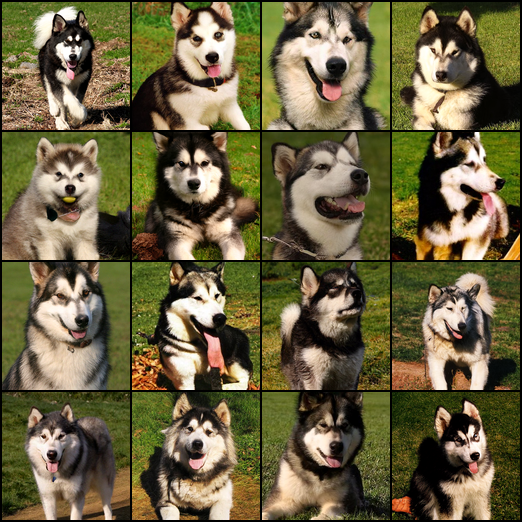}}\hfill
    \subfigure[cfg=8]{\includegraphics[width=0.23\textwidth]{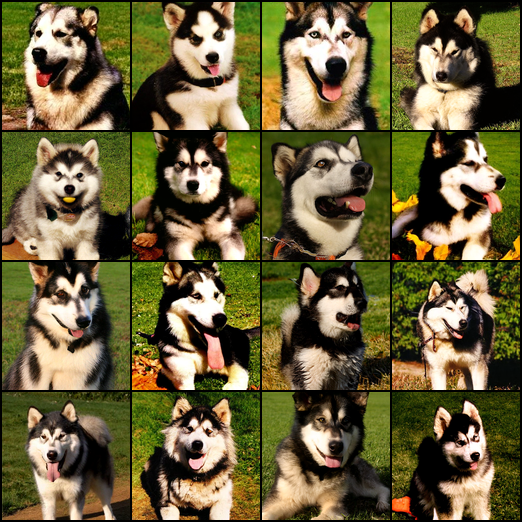}}
    \caption{Generated samples in DisCo-Diff with a cfg scale ranging from $0$ to $8$, under the class label ``malamute" on ImageNet-128. }
    \label{fig:vary-cfg}
\end{figure}

Classifier-free guidance~\cite{ho2021classifierfree}~(cfg) is a mode-seeking technique commonly used in diffusion literature, such as class-conditioned genreation~\cite{peebles2023scalable} or text-to-image generation~\cite{rombach2021highresolution}. It generally guides the sampling trajectories toward higher-density regions. We can similarly apply classifier-free guidance in the DisCo-Diff, where we treat the discrete latent as conditional inputs. We follow the convention in \cite{saharia2022photorealistic}, and the classifier-free guidance at time step $t$ is as follows:
$
    \Tilde{\mD}_\theta(\rvx, \sigma(t), \rvz) = w\mD_\theta(\rvx, \sigma(t), \rvz) + (1-w)\mD_\theta(\rvx, \sigma(t), \emptyset)
$, 
where  $\mD_\theta(\rvx, \sigma(t), \rvz)/\mD_\theta(\rvx, \sigma(t), \emptyset)$ is the conditional/unconditional models, sharing parameters. We drop the discrete latent with probability $0.1$ during training, to train the unconditional model $\mD_\theta(\rvx, \sigma(t), \emptyset)$. A mild $w$ would usually lead to improvement in sample diversity~\cite{peebles2023scalable}. Table~\ref{tab:imagenet-64-ablation} demonstrates that using a moderate guidance scale $w$=1 (we use $w=1$ and cfg=1 interchangeably in the paper) improves the FID score, suggesting that the learned discrete latent in the DisCo-Diff framework has strong indications of mode of data distribution. We further explore varying the guidance scale on ImageNet-128. As shown in Fig~\ref{fig:vary-cfg}, increasing the classifier-free guidance scale $w$ would strengthen the effect of guidance.

\section{{Algorithm Pseudocode}}
\label{app:algo}
{We provide algorithm pseudocode for the training in the first stage for denoiser and encoder~(Alg~\ref{algorithm-stage-1}) and the second stage for auto-regressive model~(Alg~\ref{algorithm-stage-2}) for clarity. We also include the pseudocode for sampling in Alg~\ref{algorithm-sampling}. Note that we generalize the equations in the main text by considering the conditional generation with condition $c$.}

\begin{algorithm}[htbp]
    \centering
    \caption{{Mini-batch training of denoiser and encoder in DisCo-Diff}}\label{algorithm-stage-1}
     \begin{algorithmic}[1]
     \STATE \textbf{Input:} Denoiser $\mD_\theta$, encoder $\mE_\phi$, training dataset $D$, batch size $\gB$, Gumbel-Softmax temperature $\tau$, training iteration $T$
     \FOR{$i=0, \dots, T-1$}
        \STATE Sample mini-batch data $\{(\rvy_i, c_i)\}_{i=1}^\gB$ from $D$
        \STATE Sample time variables $\{t_i\}_{i=1}^\gB$ from $p(t)$ and noise vectors $\{\rvn_i \sim \gN(0, \sigma_i^2 \mI)\}_{i=1}^\gB$
        \STATE Get perturbed data $\{\hat{\rvy}_i = \rvy_i + \rvn_i\}_{i=1}^\gB$
        \STATE Sample the discrete latent from the encoder $\{\rvz_i \sim \mE_\phi(\rvy_i)\}_{i=1}^\gB$ using Gumbel-Softmax relaxation with temperature $\tau$
        \STATE Calculate loss $\ell(\theta, \phi) =  \sum_{i=1}^\gB \lambda (t)\|\mD_\theta(\hat{\rvy}_i, t_i, \rvz_i, c_i)-\rvy_i\|_2^2$
        \STATE Update the network parameter $\theta$ and $\phi$ via Adam optimizer
        \ENDFOR
    \end{algorithmic}
\end{algorithm}

\begin{algorithm}[htbp]
    \centering
    \caption{{Mini-batch training of auto-regressive model in DisCo-Diff}}\label{algorithm-stage-2}
     \begin{algorithmic}[1]
     \STATE \textbf{Input:} Auto-regressive model $\mA_\psi$, encoder $\mE_\phi$, training dataset $D$, batch size $\gB$, Gumbel-Softmax temperature $\tau$, training iteration $T$
     \FOR{$i=0, \dots, T-1$}
        \STATE Sample mini-batch data $\{(\rvy_i, c_i)\}_{i=1}^\gB$ from $D$
        \STATE Sample the discrete latent from the encoder $\{\rvz_i \sim \mE_\phi(\rvy_i)\}_{i=1}^\gB$ using Gumbel-Softmax relaxation with temperature $\tau$
        \STATE Calculate loss $\ell(\psi) =  \sum_{i=1}^\gB \sum_j^m\log p_\psi((\rvz_i)_j|(\rvz_i)_{<j}, c_i)$
        \STATE Update the network parameter $\psi$ via Adam optimizer
        \ENDFOR
    \end{algorithmic}
\end{algorithm}

\begin{algorithm}[htbp]
    \centering
    \caption{{Sampling procedure of DisCo-Diff}}\label{algorithm-sampling}
     \begin{algorithmic}[1]
     \STATE \textbf{Input:} Auto-regressive model $\mA_\psi$, denoiser $\mD_\phi$, time discretization $\{t_i\}_{i=0}^{n}$, condition $c$
     \STATE \texttt{/* Sample discrete latent from auto-regressive model */}
    \FOR{$i=1, \dots, m$} 
    \STATE Sample $\rvz_i \sim p_\psi(\rvz_i|\rvz_{<i}, c)$
    \ENDFOR 
     \STATE \texttt{/* Diffusion ODE (Heun's 2nd order method) */}
     \STATE Sample $\rvx_0 \sim \gN({\bm{0}}, t_0^2\bm{I})$
     \FOR{$i=0, \dots, n-1$}
        \STATE $\rvd_i = (\rvx_i-\mD_\theta({\rvx}_i, t_i, \rvz, c))/t_i$
        \STATE $\rvx_{i+1} = \rvx_i + (t_{i+1} -t_i) \rvd_i$
        \IF{$t_{i+1} \not =0$}
        \STATE $\rvd_i' = (\rvx_{i+1}-\mD_\theta({\rvx}_{i+1}, t_{i+1}, \rvz, c))/t_{i+1}$
        \STATE $\rvx_{i+1} = \rvx_i + (t_{i+1} -t_i) (\frac12\rvd_i+\frac12 \rvd_i')$
        \ENDIF
        \ENDFOR
    \end{algorithmic}
\end{algorithm}

\section{ImageNet Experiments} \label{app:imagenet_exps_details}

\subsection{Architecture}

\begin{table}[htbp]
    \centering
    \begin{tabular}{c c c c}
    \toprule
         &  Feature maps  & Attention resolution & Encoder patch size\\
        \midrule
        ImageNet-64 & 1-2-3-4 ($\times {192}$) & (8,16,32) &  $8\times 8$\\
        \midrule
        ImageNet-128 & 1-2-4-16 ($\times {128}$) & (16,32,64) & $16\times 16$\\
        \bottomrule
    \end{tabular}
    \caption{Specific network configurations on ImageNet}
    \label{tab:imagenet-conf}
\end{table}

For all the ImageNet experiments, we fix the latent dimension $m=10$ in DisCo-Diff, and the codebook size for each discrete latent to $1000$. Below we provide architecture details for the denoiser network, encoder, and auto-regressive model. Table~\ref{tab:imagenet-conf} also lists some key network configurations. Please see the source code in the Supplementary Material for all low-level details.

\paragraph{Denoiser Neural Network.} (1) \textbf{ImageNet-64}: We use the same UNet architecture in EDM~\cite{Karras2022ElucidatingTD} for ImageNet-64, with newly injected cross-attention layers after each self-attention layers in each residual block. We feed the discrete vector into a six-layer Transformer encoder (with latent dimension 192) to obtain the corresponding embeddings for discrete variables. These embeddings are then input into the cross-attention layers. (2) \textbf{ImageNet-128}: We employ the UViT design in simple diffusion~\cite{hoogeboom2023simplediffusion} and VDM++~\cite{Kingma2023UnderstandingDO}. UViT uses convolutional layers for down-/up-sampling, and a 36-layer ViT to process the lowest-resolution feature maps in the bottleneck, to strike a better balance between expressiveness and computation. Since the authors didn't release the code and model, we reimplemented the architecture by ourselves. We empirically observe that the convolutional blocks in EDM work better than the ones described in the simple diffusion paper, so we combine the up-/down-sampling blocks in EDM with the 36-layer ViT at the bottleneck layer. We further introduce a cross-attention layer for discrete latent in each up-/down-sampling block, and every three Transformer blocks in the ViT (\textit{e.g.,} 12 new cross-attention layers in the ViT) to save computation.
\paragraph{Encoder.} We utilize a 12-layer standard ViT~\cite{dosovitskiy2021vit} as the backbone for encoder. Its latent dimension is 768 and the number of attention heads is 12. The patch size for ImageNet-64 is $8\times 8$ and for ImageNet-128 is $16\times 16$. We treat each of the $m$ discrete latents as a classification token, and concatenate their embeddings with the path embeddings.
\paragraph{Auto-regressive model.} We use a standard Transformer decoder~\cite{Vaswani2017AttentionIA}, with a depth of $12$, a number of heads of $8$, and a latent dimension of $512$. The inference time of the auto-regressive model is much smaller than the iterative denoising process, given that the discrete latent only has $10$ dimensions. To generate 32 images on ImageNet-128, the auto-regressive model takes 0.44 seconds, while the diffusion model takes 78 seconds for 114 NFE, with an average of 0.68 s/NFE on a single NVIDIA A100 GPU.

\subsection{Training and Sampling}

We borrow the preconditioning techniques, training noisy schedule, optimizers, exponential moving average~(EMA) schedule, and hyper-parameters from previous state-of-the-art diffusion model EDM~\cite{Karras2022ElucidatingTD} on ImageNet-64. We employ the shifted EDM-monotonic noisy schedule proposed in VDM++~\cite{Kingma2023UnderstandingDO} on ImageNet-128, and keep other training details the same in ImageNet-64. We use the Gumbel-Softmax~\cite{Jang2016CategoricalRW} as the continuous relaxation for the discrete latents. The temperature $\tau$ in Gumbel-Softmax controls the smoothness of the categorical distribution. When $\tau \to 0$, the expected value of the Gumbel-Softmax is the same as the one of the underlying predicted distribution. As we increase $t$, the Gumbel-Softmax would gradually converge to a uniform distribution. Hence, a relatively large $\tau$ effectively provides regualization effects. During training, we set the $\tau$ to a constant $1$. However, for the extraction of latents from training images, which aids in constructing the dataset for the second stage of the auto-regressive model, we adjust $\tau$ to a lower value of 0.01.

We use Heun's second-order ODE solver as the default ODE sampler, which is proven effective in previous works~\cite{Karras2022ElucidatingTD, xu2023pfgmpp}. We directly use the hyper-parameters for the 623 NFE setting in Restart sampler~\cite{Xu2023RestartSF} on ImageNet-64, for DisCo-Diff. 

\subsection{Evaluation}

For the evaluation, we follow the standard protocol and compute the Fréchet distance between 50000 generated samples and the training images.

\section{Molecular Docking} \label{app:docking_general}

In this appendix, we will introduce the task of molecular docking and some of the existing approaches to tackle it for readers who are not familiar with this field. Experienced readers may skip to Section \ref{app:docking_latent} where we start describing the details of our docking approach.

\subsection{Task Overview} 

Molecular docking consists of finding the 3D structure that a protein (also referred to as receptor) and a small molecule (or ligand) take when binding. This is an important task in drug design because most drugs are small molecules that operate by binding to a specific protein of interest in our body and inhibiting or enhancing its function. The common particular instantiation of the docking problem that we consider is also referred to as rigid blind docking i.e. where we are given as input the correct protein structure (\textit{rigid}) but are not provided any information about where the ligand will bind on the protein nor the conformations it will take (\textit{blind}). 

Ground truth data for training and testing is obtained through experimental methods, like X-ray crystallography, that, for each protein-ligand complex, allow to observe a particular pose that protein and ligand took when binding together inside of the crystal. Although there may be other poses that this particular protein and ligand may take when binding in a natural environment, methods are evaluated based on their capacity to retrieve the crystal pose. This accuracy is typically computed as the percentage of test complexes where the predicted structure of the ligand (also referred to as ligand pose) is within a root mean square distance (RMSD) of 2\AA{} from the ground truth when aligning the protein structures.

\subsection{Related Work}

Traditional approaches tackled the task via a search-based paradigm where, given a scoring function, they would search over possible ligand poses with an optimization algorithm to find a global minimum \cite{halgren2004glide, trott2010autodock}. Recently, deep learning methods have been trying to speed up this search process by generating poses directly through a neural network. Initial approaches \cite{equibind, Lu2022TankBind} used regression-based frameworks to predict the pose, but, although significantly faster, they did not outperform traditional methods.

\citet{corso2023diffdock} argued that the issue with these regression-based approaches is their treatment of the uncertainty in the multimodal model posterior pose distribution. They also proposed DiffDock, a diffusion-based generative model to generate docked poses that was able to outperform previous methods, which we use as a starting point for the integration of our DisCo-Diff approach to diffusion.

Most deep learning approaches to docking model the data as a geometric graph or point cloud in 3D. The nodes of this graph are the (heavy) atoms of the ligand and, typically, the C-alpha carbon atoms of the protein backbone (sometimes full-atom representations are also used for the protein but these are less common for computational complexity reasons). These nodes are connected by edges in case of chemical bonds or pairwise distances below a certain cutoff. Neural architectures learn features over the nodes of this graph through a number of message passing layers, the geometric structure is encoded via invariant (e.g. relying only on distance embeddings, see \citet{schutt2017schnet}) or equivariant operations \cite{geiger2022e3nn}.

\subsection{Latent Variables} \label{app:docking_latent}

We design each latent variable to take values indicating one of the nodes in the protein-ligand joint graph. Therefore the codebook size for the latent variable of any given protein-ligand complex is equal to the total number of nodes in the graph i.e. the sum of the number of atoms in the ligand and the number of residues in the protein. With this choice, intuitively each latent variable will indicate one particular atom or residue involved in some key component of the protein-ligand interaction. For example, using two latents the model can learn to indicate a geometric contact between a pair of nodes in the final representation. Further note, each "codebook", when considered as a set of one-hot vectors indicating notes, has a permutation equivariance property with the nodes of the graph (because they are associated with node properties): if the nodes of the input graph are permuted each latent variable coming out of the encoder or autoregressive model will also have its codebook representation permuted. 

\subsection{Architecture Details} \label{app:docking_architecture}

\textbf{Denoiser. } This design choice for the discrete latents codebooks fits very well with the preexisting DiffDock's denoiser architecture composed of equivariant message passing layers \cite{geiger2022e3nn}. Each latent variable is encoded in a binary label for each node which is set to zero for all nodes except the one indicated by the latent. These binary labels are concatenated to the initial node features while the rest of the denoiser is kept unchanged. With probability 0.1 during training we drop the latents, in this case, the binary labels are set to zero for all latents, and a learnable null-embedding is fed to all initial node features. 

\textbf{Encoder. } The encoder and autoregressive models adopt very similar architectures to the denoiser with a few key distinctions. The encoder takes as input the ground truth pose of the ligand, learns features for each node through message passing, and finally \textit{m} separate feedforward MLPs (where \textit{m} is the number of latents) with a one-dimensional output are applied to each node representation. The concatenated outputs of each of these MLPs form the logit vectors for each of the latent variables which are passed through the Gumbel-Softmax discretization step.

\textbf{Autoregressive model. } Unlike the image synthesis experiments setting where the images are often generated with relatively vague conditioning information, for docking, we are interested in generating ligand poses conditioned on a particular protein (structure) and ligand. This conditioning information significantly influences the posterior pose distribution and consequently the learned latent variables. Therefore, we need to condition the autoregressive model on the protein structure and the ligand. We achieve this, once again, through an equivariant message passing network, operating on an input composed of the protein structure and the ligand. The latter is centered at the protein's center, given an arbitrary conformer (i.e. molecular conformation) from RDKit \cite{landrum2013rdkit} and a uniformly random orientation. Like the denoiser, the autoregressive model takes as input the additional binary node labels for the existing latents (masked out appropriately during training), and, like the encoder, it uses its final node embeddings to predict the logits for the next latent variable.

\subsection{Experimental Details} \label{app:docking_exps_details}

For the docking experiments, we follow the datasets and procedures established by \citet{equibind} and \citet{corso2023diffdock}. Data for training and evaluation comes from the PDBBind dataset \cite{liu2017PDBBind} with time-based splits (complexes before 2019 for training and validation, selected complexes from 2019 for testing). 

\textbf{Denoiser.} We use a denoiser architecture analogous to the one proposed by \citet{corso2024discovery}, which is a smaller version of DiffDock's original architecture where the main changes are: (1) 5 convolutional layers (vs 6 of the original DiffDock's architecture) (2) node representations with 24 scalars and pseudoscalars and 6 vectors and pseudovectors (vs respectively 48 and 10) (3) spherical harmonics order limited to 1 (vs 2). These changes, although somewhat affecting the inference quality, make training and testing of the models significantly more affordable (from 18 days on 4 GPUs to 9 days on 2 GPUs for training).

We keep the same denoiser architecture for both the baseline without discrete latents (DiffDock-S) and our model and apply similar hyperparameter searches when applicable to both models. At inference time, similarly to \citet{corso2023diffdock}, we take 40 independent samples and use the original DiffDock's confidence model\footnote{The confidence model is an additional model, \citet{corso2023diffdock} trained to select the most likely correct poses out of the diffusion models samples. The reader can think of this as trying to select the maximum likelihood pose.} to select the top one. For DisCo-DiffDock each of the samples is taken by independently sampling from the autoregressive model and then the (conditioned) denoiser.

\textbf{Encoder.} For the encoder, we use a similar but slightly smaller architecture with 3 convolutional layers, 24 scalars, and 4 vectors. We set the number of discrete latent variables (each taking values over the whole set of possible nodes in the joint graph) to two, as we found this to equilibrate the complexity of the generative task between the score and autoregressive models. 

\textbf{Autoregressive model.} One challenge with the autoregressive model in this domain is its tendency to overfit the latent variables in the training set given the limited training data, the complexity of the conditioning information, and the low training signal that discrete labels provide. Therefore we found it beneficial to design the autoregressive model to use the pretrained layers of the denoiser itself. In particular, we simply add independent MLPs for each latent variable that are applied to the final scalar representations of the nodes. During the autoregressive training, for the first five epochs, the weights of the convolutional layers are frozen. 

\textbf{Inference hyperparameters.} For inference, we maintain the number of inference steps from DiffDock (20) and, for both DisCo-DiffDock and the baseline, we tune on the validation set the sampling temperature for the different components of the diffusion similarly to how it was done by \citet{ketata2023diffdockpp}. For DisCo-DiffDock we also tune the temperature used to sample the autoregressive model. We find, with 40 samples, to be beneficial to set this temperature $>1$ while the diffusion sampling temperature $<1$, this corresponds to encouraging exploration of different binding modes while trying to obtain the maximum likelihood pose for each mode. This further highlights the advantage provided by enabling the decomposition of different degrees of uncertainty. Please see the source code in the Supplementary Material for all low level details and hyperparameters used.

\section{Gaussian Mixture Experiments} \label{app:exps_details}


\subsection{Data Generation}
For the toy example in section~\ref{sec:method}, we set the true data distribution to a mixture of eight Gaussian components:
\begin{align*}
    p_{data}(\rvx) = \frac18 \sum_{i=1}^8 \gN(\rvx; \bm{\mu}_i, \sigma_i\bm{I}_{2\times 2})
\end{align*}
where $\forall i, \sigma_i = 0.2$, and 
\begin{align*}
    \bm{\mu}_1 = \begin{pmatrix}3\\0\end{pmatrix}, \quad \bm{\mu}_2 = \begin{pmatrix}-3\\0\end{pmatrix},\quad \bm{\mu}_3 = \begin{pmatrix}0\\3\end{pmatrix},\quad \bm{\mu}_4 = \begin{pmatrix}0\\-3\end{pmatrix},\\
     \bm{\mu}_5 = \begin{pmatrix}\frac{3}{\sqrt{2}}\\\frac{3}{\sqrt{2}}\end{pmatrix}, \quad \bm{\mu}_6 = \begin{pmatrix}\frac{3}{\sqrt{2}}\\\frac{-3}{\sqrt{2}}\end{pmatrix},\quad \bm{\mu}_7 = \begin{pmatrix}\frac{3}{\sqrt{2}}\\\frac{-3}{\sqrt{2}}\end{pmatrix},\quad \bm{\mu}_8 = \begin{pmatrix}\frac{-3}{\sqrt{2}}\\\frac{-3}{\sqrt{2}}\end{pmatrix}.
\end{align*}

To construct the toy dataset, we randomly sampled 1000 data points from each component, totaling 8000 data points. We visualize the KDE plot of the generated data in Fig.~\ref{fig:toy-vis}(a).

\subsection{Training and Sampling}

We employ a four-layer MLP as the diffusion decoder (Denoiser Neural Network) $\gG$, for both Disco-Diff and diffusion models. We use a three-layer MLP as the encoder $\gE$ in Disco-Diff. We set the latent dimension of discrete latent to 1 and the vocabulary size to 8. Ideally, each discrete latent should correspond to a Gaussian component, and the time-dependent scores for a single Gaussian component have a simple analytical expression. We leverage this simplicity and reparameterize the output of diffusion decoder as $\gG(\rvx, t, z) = \frac{\gF(z) - \rvx} { t^2+\sigma_1^2} + \gH(\rvx, t)$, where $\gF$ is the embedding for each discrete latent $z$ and $\gH$ is a four-layer MLP. The model optimization uses the Adam optimizer with a learning rate of 1e-3.

For sampling, we use the Heun's second-order sampler. We followed the time discretization scheme in EDM~\cite{Karras2022ElucidatingTD} with 50 sampling steps.

\subsection{Metric}

We detail the metrics used in Fig.~\ref{fig:toy-analysis}. 
The curvature for points $\rvx(t)$ on ODE trajectory $d\rvx / dt = \gG(\rvx, t, z)$ ($z$ is null in diffusion models) is defined as $\kappa(\rvx(t)) = \frac{||\partial_t \bm{T}(\rvx(t), t)||}{||\rvx'(t)||}$ where $\bm{T}(\rvx, t) = \frac{\gG(\rvx(t), t, z)}{||\gG(\rvx(t), t, z)||}$ is the unit tangent vector. We can approximate the curvature by finite difference: $\kappa(\rvx(t)) = \frac{||\partial_t \bm{T}(t)||}{||\rvx'(t)||} \approx \frac{||\bm{T}(t)-\bm{T}(t-\Delta t)||}{||\rvx(t)-\rvx(t-\Delta t)||}$. We approximate $\rvx(t-\Delta t)$ by a single Euler step, \textit{i.e.,} $\rvx(t-\Delta t) = \rvx(t) -  \gG(\rvx, t, z)\Delta t$. In Fig.~\ref{fig:toy-analysis}(a), we report the expected curvature given the backward time when simulating the ODE, \textit{i.e.,} $\E_{\rvx(t)}\left[\frac{||\bm{T}(t)-\bm{T}(t-\Delta t)||}{||\rvx(t)-\rvx(t-\Delta t)||}\right]$. We set the time elapsed to $\Delta t = 0.001$.

In Fig.~\ref{fig:toy-analysis}(b), we measure the complexity of the trained neural networks using the expected squared Frobenius norm of the network's Jacobians, \textit{i.e.,} $\E_{\rvx(t)}\left[{||\nabla_\rvx \gG(\rvx, t, z)||_F^2}\right]$.

Additionally, to quantitatively evaluate the generation quality, we report the Wasserstein-2~(W-2) distance between the generated distribution and the ground truth distribution. In the DisCo-Diff model, the W-2 distance is at 0.118, compared to 0.27 in the standard diffusion model. It suggests that DisCo-Diff better captures the multimodal distribution, even in 2-dim space.




\begin{figure}[tbp]
    \centering
    \subfigure[ImageNet-64]{\includegraphics[width=0.35\textwidth]{figures/loss_64.pdf}}
    \subfigure[ImageNet-128]{\includegraphics[width=0.35\textwidth]{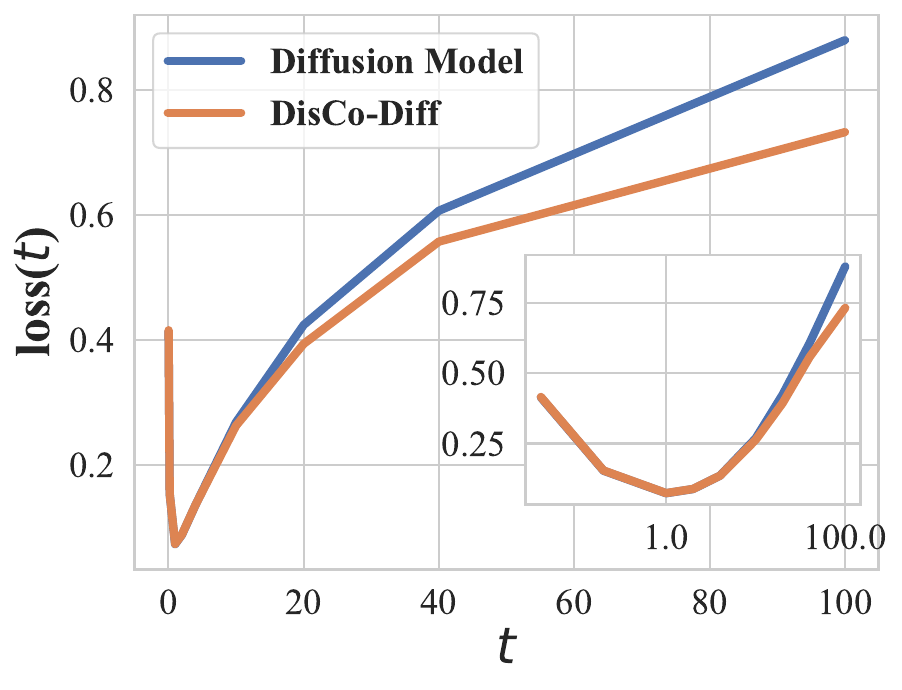}}
    \caption{Averaged training loss versus noise level $t$.}
    \label{fig:loss-app}
\end{figure}

\section{Additional Samples and Experiments}
\label{app:extra_sample}

\subsection{Loss Analysis}

In Fig.~\ref{fig:loss-app}, we provide the loss versus time curve on both ImageNet-64 and ImageNet-128 datasets. We have also included a log-scale version of the x-axis in the inset plot.

\subsection{Class-conditoned ImageNet-64}

\begin{figure}
    \centering
    \includegraphics[width=0.7\textwidth]{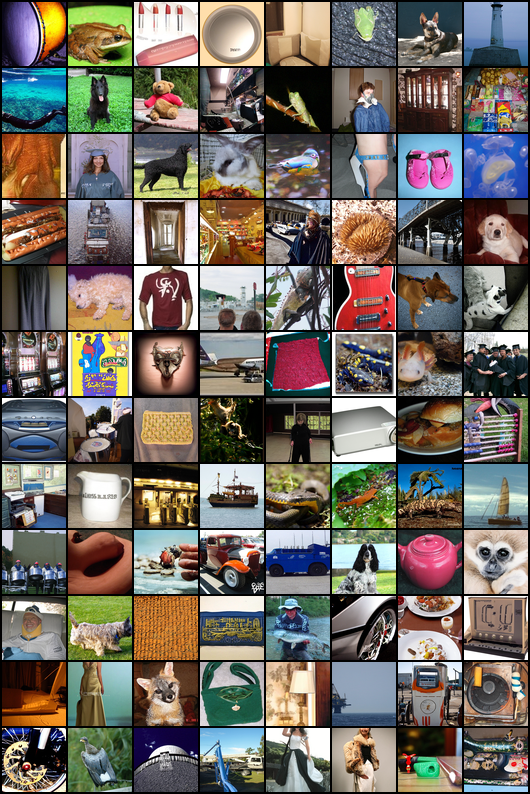}
    \caption{Generated samples by DisCo-Diff on class-conditioned ImageNet-64, with ODE sampler (FID=1.65, NFE=78).}
    \label{fig:app-imagenet64}
\end{figure}

We provide extended samples generated by DisCo-Diff in Fig.~\ref{fig:app-imagenet64}.

\subsection{Class-conditoned ImageNet-128}
\label{app:imagenet-128}
\begin{figure}
    \centering
    \includegraphics[width=0.8\textwidth]{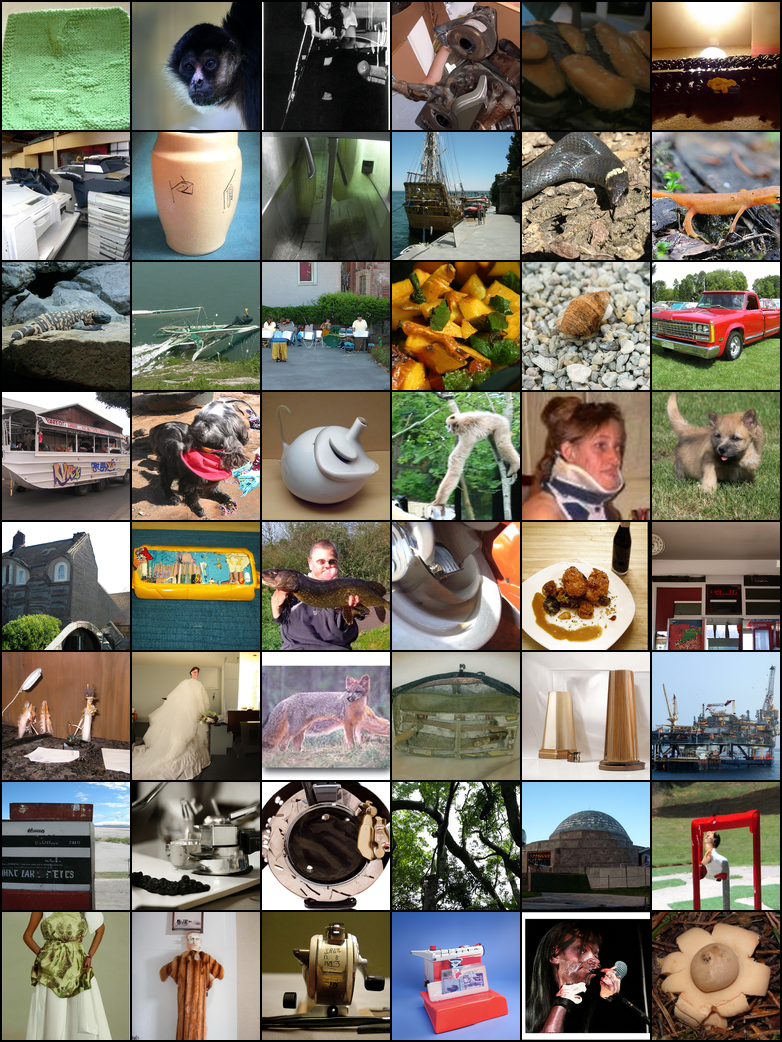}
    \caption{Generated samples by DisCo-Diff on class-conditioned ImageNet-128, with ODE sampler (FID=2.08, NFE=114).}
    \label{fig:app-imagenet128}
\end{figure}

\begin{figure}
    \centering
    \subfigure[DisCo-Diff ODE, FID=1.98]{\includegraphics[width=0.32\textwidth]{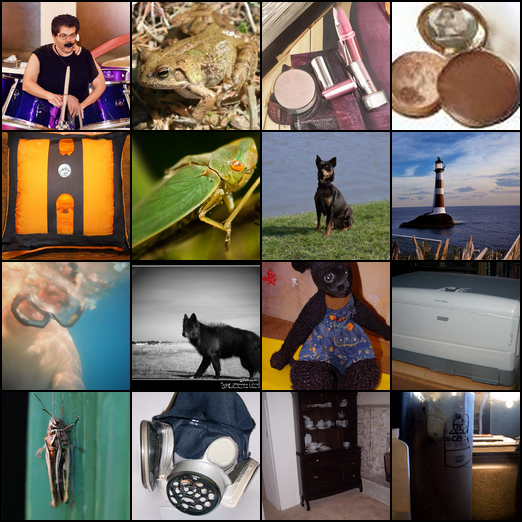}}\hfill
    \subfigure[DisCo-Diff ODE~($t\ge 10$) and DisCo-Diff DDPM~($t<10$), , FID=2.78]{\includegraphics[width=0.32\textwidth]{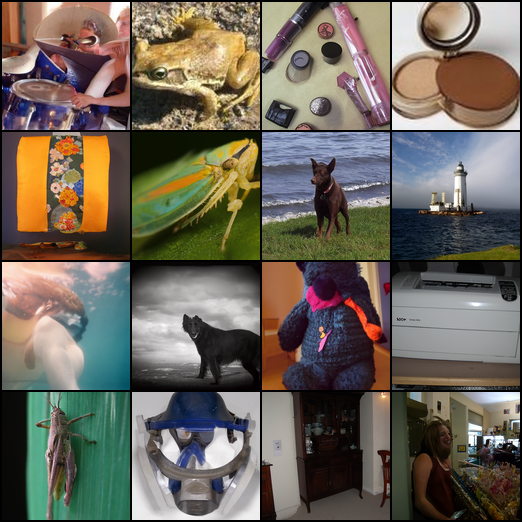}}\hfill
    \subfigure[DisCo-Diff ODE~($t\ge 10$) and VDM DDPM~($t<10$), FID=1.73]{\includegraphics[width=0.32\textwidth]{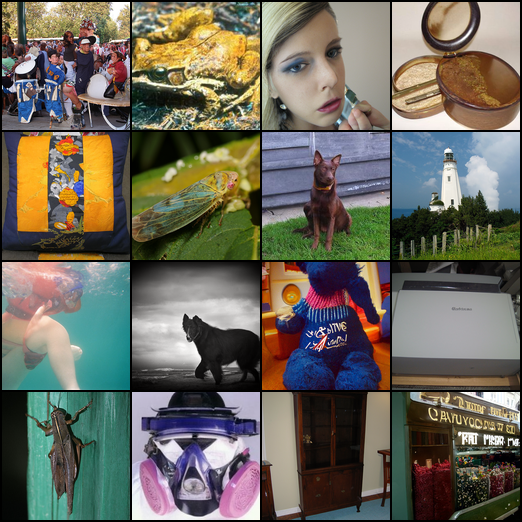}}
    \caption{Images generated by different samplers. When using the DDPM sampler at smaller times~(b), the generated images exhibit a slight over-smoothing issue, losing some high-frequency details in comparison to those produced with the VDM++ DDPM sampler at smaller times. Note that FID score typically penalizes over-smooth samples. This observation supports our hypothesis that the use of the DDPM sampler in DisCo-Diff at smaller times can in certain situations overlook high-frequency details.}
    \label{fig:vis-mix}
\end{figure}

\begin{figure}
    \centering
    \includegraphics[width=0.8\textwidth]{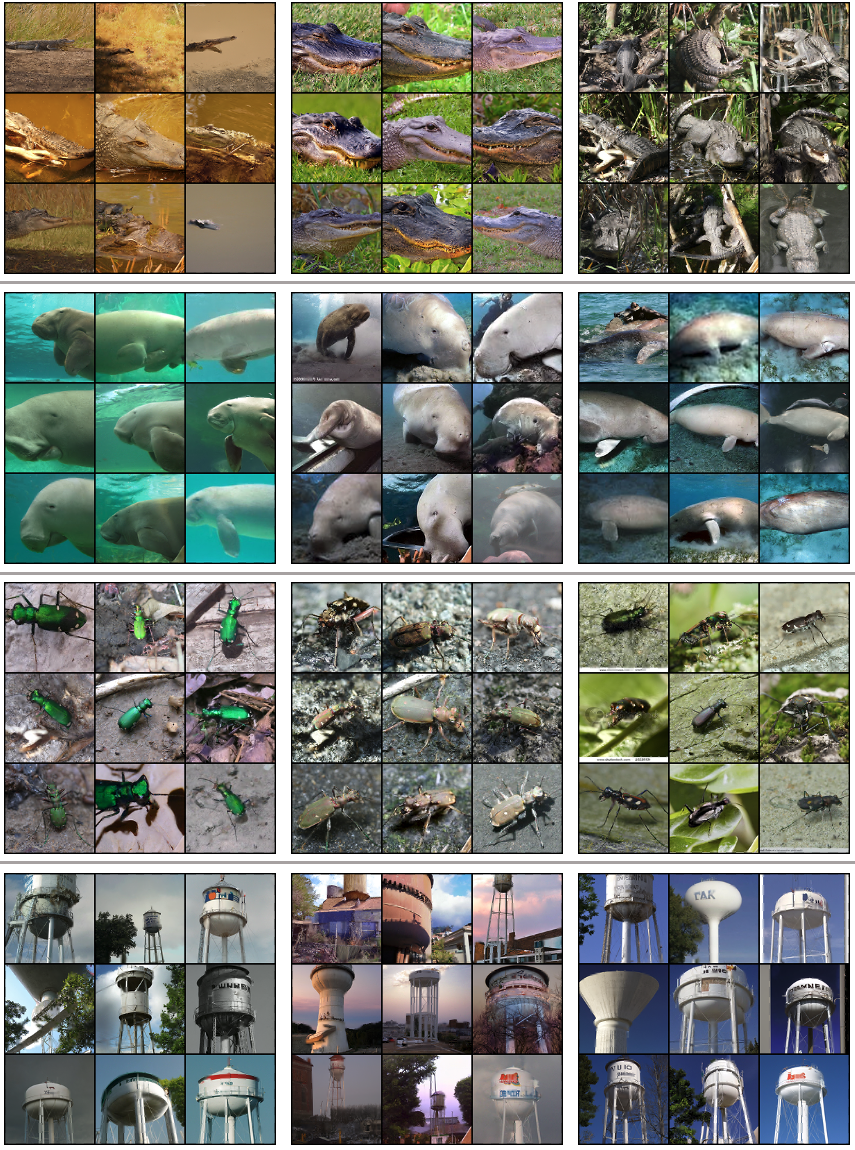}
    \caption{Generated samples by DisCo-Diff on class-conditioned ImageNet-128, with ODE sampler. Samples in each grid share the same latent, and grids in each row share the same class labels. We can see that generally, images sharing the same discrete latents demonstrate similar global characteristics, such as shape, layout, and color, despite being under the same class. It suggests that discrete latents provide complementary information to the class labels.}
    \label{fig:app-imagenet128-z}
\end{figure}

We provide extended samples generated by DisCo-Diff in Fig.~\ref{fig:app-imagenet128}. We also visualize samples with shared discrete latents in Fig.~\ref{fig:app-imagenet128-z}. We further provide samples using different samplers in Fig.~\ref{fig:vis-mix}, to highlight the over-smoothing issue of DisCo-Diff at smaller times we observed for this model when using the DDPM sampler. Although, in theory, the VDM++ model and DisCo-Diff should learn the same field at small times, in practice, we observe that VDM++ DDPM provides samplers with richer details compared to DisCo-Diff DDPM at smaller times. It supports our hypothesis that the discrete latents tend to divert the model to overlook the high-level details on ImageNet-128, when using the DDPM sampler and the network architecture in \cite{Kingma2023UnderstandingDO}. We would like to emphasize that we only observed this behavior for this one model and dataset. In all other experiments, discrete latents universally improved performance for all stochastic and non-stochastic samplers, even when used for all times $t$.

\subsection{Group Hierarchical DisCo-Diff}

\begin{figure}
    \centering
    \includegraphics[width=0.8\textwidth]{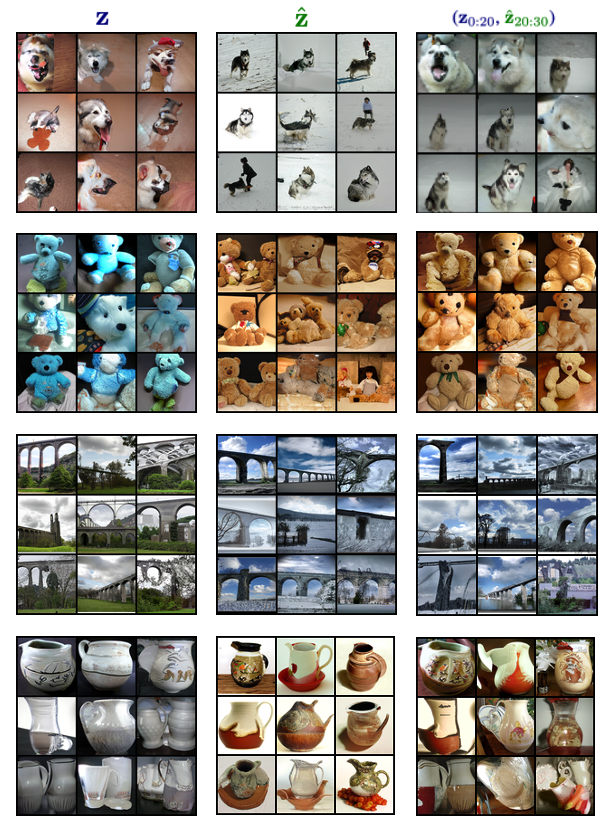}
    \caption{Generated images with a shared latent, using group hierarchical DisCo-Diff trained on ImageNet-64. \textit{Left:} Shared latent {\color{black!50!blue}{$\rvz$}}. \textit{Middle:} Shared latent {\color{black!50!green}{$\hat{\rvz}$}}. \textit{Right:} Shared latent ({\color{black!50!blue}{$\rvz_{0:20}$}}, {\color{black!50!green}{$\hat{\rvz}_{20:30}$}}), where the first 20 coordinates are from {\color{black!50!blue}{$\rvz$}} and the last 10 coordinates are from {\color{black!50!green}{$\hat{\rvz}$}}. We can see that the generated images from composed latents generally inherit the shape from images generated by {\color{black!50!blue}{$\rvz$}}, and the color from images generated by {\color{black!50!green}{$\hat{\rvz}$}}. }
    \label{fig:app-hier-comp}
\end{figure}

We further provide extended samples from the Group hierarchical DisCo-Diff. Fig.~\ref{fig:app-hier-comp} showcases the generated images when composing two discrete latents together, \textit{i.e.,} ({\color{black!50!blue}{$\rvz_{0:20}$}}, {\color{black!50!green}{$\hat{\rvz}_{20:30}$}}). We can see that the generated images from composed latent generally inherit the shape from images generated by {\color{black!50!blue}{$\rvz$}}, and the color from images generated by {\color{black!50!green}{$\hat{\rvz}$}}. 

\begin{figure}
    \centering
    \includegraphics[width=0.8\textwidth]{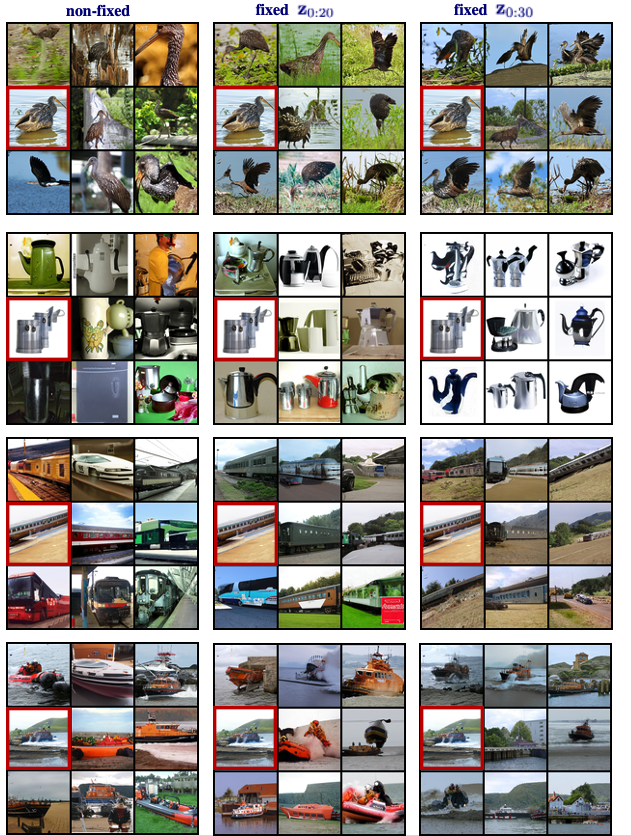}
    \caption{Progressively fixing more subcoordinates of the discrete latents, using our group hierarchical DisCo-Diff on ImageNet-64. \textit{Left:} Randomly sampled {\color{black!50!blue}{$\rvz$}}. \textit{Middle:} Fixing the first 20 coordinates {\color{black!50!blue}{$\rvz_{:20}$}} as the one derived from the red-boxed image, sampling the rest. \textit{Right:} Fixing the whole 30-dim. {\color{black!50!blue}{$\rvz$}} as the one derived from the red-boxed image. The figure shows the effect when progressively fixing more coordinates of the discrete latent, and sampling the remaining coordinates by the auto-regressive model. The images first converge in shape/layout, and subsequently converge in color/texture.}
    \label{fig:app-hier-auto}
\end{figure}

Fig.~\ref{fig:app-hier-auto} further shows the effect when progressively fixing more coordinates of the discrete latent, and sampling the remaining coordinates by the auto-regressive model. The images first converge in shape/layout, and subsequently converge in color/texture.

\subsection{Overfitting and Encoder Collapse in Continuous Latent}

\begin{figure}[htbp]
    \centering
    \subfigure[Real image]{\includegraphics[width=0.15\textwidth]{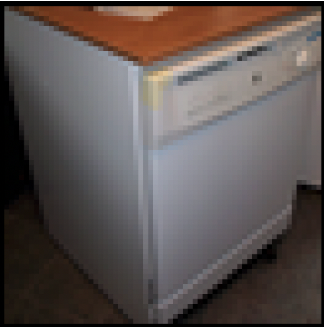}}\hfill
    \subfigure[KLD weight=$0.1$]{\includegraphics[width=0.23\textwidth]{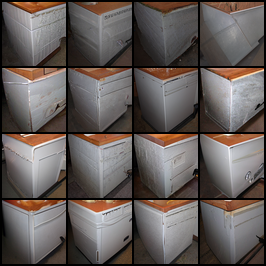}}\hfill
    \subfigure[KLD weight=$1$]{\includegraphics[width=0.23\textwidth]{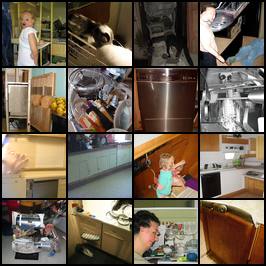}}\hfill
    \subfigure[DisCo-Diff]{\includegraphics[width=0.23\textwidth]{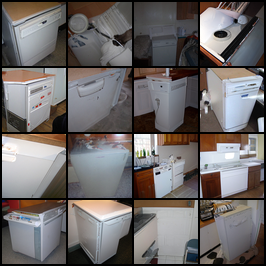}}
    \caption{We derived the continuous / discrete latents from the real image in (a) and fed the latents into the denoisers. (b): The continuous latent model exhibits the overfitting issue when KLD weight equals to 0.1, as all the generated images are very similar to the real image. (c): When applying a stronger KL regulation (KLD weight=$1$) on the continuous latent, the model exhibits encoder collapse -- the denoiser would ignore the continuous latent. (d):  DisCo-Diff generated a batch of diverse samples, sharing a similar high-level layout and color with the real image. }
    \label{fig:overfit}
\end{figure}

In this section, we show that it is difficult to control the amount of information stored in the continuous latents, which would lead to either overfitting or encoder collapse. To illustrate the issue, we derived the continuous / discrete latents from a specific real image, and fed the corresponding latents into the denoisers. As shown in \Figref{fig:overfit}, when KLD weight=$0.1$, the continuous latent model exhibits the overfitting issue, as all the generated images are very similar to the training image. It also indicates that the encoder squeezes excessive information in the continuous latent when using a smaller KLD weight, which complicates generative training in the second stage. When KLD weight=$1$, the model exhibits encoder collapse -- the denoiser would ignore the continuous latent. We observe that even using a different continuous latent, the model will still generate an identical batch of samples. In contrast, DisCo-Diff generated a batch of diverse samples, sharing a similar high-level layout and color with the training image. This indicates that the discrete latents in DisCo-Diff encode global layout and color attributes --- key statistical elements crucial for the diffusion process in Euclidean space. This aligns with more direct and straighter ODE trajectories.

\end{document}